%% file: main_revised_R1.tex
\documentclass[fleqn,10pt]{wlscirep}
\usepackage[utf8]{inputenc}
\usepackage[T1]{fontenc}
\usepackage[table]{xcolor}

\usepackage{amsmath}
\usepackage{amsthm}
 \usepackage{array}
\usepackage{amssymb}
\usepackage{mathtools}
\usepackage{bm}
\usepackage{bigints}

\newcommand{\R}{\mathbb{R}}
\newcommand{\mH}{\mathcal H}
\newcommand{\bx}{\mathbf x}
\newcommand{\vv}{\mathbf v}
\newcommand{\vs}{\mathbf s}
\newcommand{\rd}{\mathrm d}

\def \rd {{\rm d}}

\def\E{{\mathbb E}}

\def\0{{\mathbb 0}}

\def\R{{\mathbb R}}

\def\bE{{\mathbb E}}
\def\bR{{\mathbb R}}
\def\bx{{\mathbf x}}
\def\mB{{\mathcal B}}
\def\tb{{\mathbf{b}}}

\def\rd{{\mathrm{d}}}

\newtheorem{theorem}{Theorem}
\newtheorem{definition}[theorem]{Definition}

\newtheorem{lemma}[theorem]{Lemma}

\newtheorem{assumption}{Assumption}

\title{FunDiff: Diffusion models over function spaces for physics-informed generative modeling}

\author[1,+]{Sifan  Wang}
\author[2,+]{Zehao Dou}
\author[2]{Siming Shan}
\author[3]{Tong-Rui Liu}
\author[2,4,*]{Lu Lu}
\affil[1]{Institute for Foundations of Data Science, Yale University, New Haven, CT 06511, USA}
\affil[2]{Department of Statistics and Data Science, Yale University, New Haven, CT 06511, USA}
\affil[3]{Laboratoire Navier, École nationale des ponts et chaussées, Institut Polytechnique de Paris, University Gustave Eiffel, CNRS UMR 8205, Champs-sur-Marne, 77420, France}
\affil[4]{Department of Chemical and Environmental Engineering, Yale University, New Haven, CT 06511, USA}

\affil[*]{Corresponding author. Email: lu.lu@yale.edu}
\affil[+]{These authors contributed equally to this work.}

\begin{abstract}
Recent advances in generative modeling---particularly diffusion models and flow matching---have achieved remarkable success in synthesizing discrete data such as images and videos. However, adapting these models to physical applications remains challenging, as the quantities of interest are continuous functions governed by complex physical laws.
To address this, we introduce \textbf{FunDiff}, an efficient and robust framework for generative modeling in function spaces. FunDiff combines a latent diffusion process with a function autoencoder architecture to handle input functions with varying discretizations, generates continuous functions that can be evaluated at arbitrary locations, and seamlessly incorporate physical priors.  These priors are enforced through architectural constraints or physics-informed loss functions, ensuring that generated samples satisfy fundamental physical laws.
We theoretically establish minimax optimality guarantees for density estimation in function spaces, demonstrating that diffusion-based estimators achieve optimal convergence rates under suitable regularity conditions.
We further demonstrate the practical effectiveness of FunDiff across diverse applications in fluid dynamics and solid mechanics. Empirical results indicate that our method can generate physically consistent samples with high fidelity to the target distribution, and exhibit robustness to noisy and low-resolution data. 
\end{abstract}

\begin{document}

\flushbottom
\maketitle
% * <john.hammersley@gmail.com> 2015-02-09T12:07:31.197Z:
%
%  Click the title above to edit the author information and abstract
%
\thispagestyle{empty}

\section*{Introduction}
Generative modeling, a cornerstone of modern machine learning, aims to learning and sampling complex data distributions $P(x)$ by constructing models that can generate new instances consistent with the learned distribution \cite{kingma2013auto, goodfellow2020generative}. Among recent advances in this domain, diffusion models and flow matching have emerged as powerful frameworks that achieve state-of-the-art performance. Diffusion models \cite{ho2020denoising, song2020score} learn to reverse a forward noising process by gradually denoising data through a Markov chain, which iteratively transforms random noise into high-quality samples. In parallel, flow matching \cite{lipman2022flow, liu2022flow} provides an alternative approach by learning continuous normalizing flows via optimal transport theory, directly modeling probability paths that transform a tractable prior distribution into the target data distribution.

Diffusion models have shown remarkable success across a wide range of applications, including high-resolution image synthesis \cite{saharia2022photorealistic, dosovitskiy2020image, jaegle2021perceiver}, audio generation \cite{kong2020diffwave}, protein structure prediction \cite{wu2024protein}, and molecular design \cite{xu2022geodiff}. 
{\color{black} Despite these advances, the focus has largely remained on discrete data types, such as images, text, and molecular graphs, whereas many scientific domains, especially in physics, demand the generation of continuous function data, where each sample represents a continuous function (e.g., a velocity or temperature field) defined over spatial and temporal domains.}
For instance, generating velocity fields in turbulent fluid dynamics, simulating deformation patterns in complex materials under external loadings, and forecasting temperature variations in climate models all require handling functions defined over continuous spatial and temporal domains. These problems inherently involve infinite-dimensional function spaces, a challenge that recent advances in operator learning frameworks have begun to address effectively \cite{li2021fourier,lu2021learning,jin2022mionet,lu2022comprehensive,jiao2021one,wu2025coast,zhang2024federated}. This presents unique challenges: 
\begin{center}
   {\color{black} \textit{How can we extend generative modeling to generate continuous functions that can be evaluated at arbitrary points, while preserving important structural or physical properties inherent in scientific problems?}}
\end{center}

{\color{black} Prior work in computer vision and graphics has extensively explored the concept of function-space generation through implicit neural representations and generative models, such as DeepSDF\cite{park2019deepsdf}, NeRF\cite{mildenhall2021nerf}, From Data to Functa\cite{dupont2022data}, and SIREN\cite{sitzmann2020implicit}, as well as their diffusion-based extensions including 3DShape2VecSet\cite{zhang20233dshape2vecset}, Shap-E\cite{jun2023shap}, and HyperDiffusion\cite{erkocc2023hyperdiffusion}. 
These methods treat signals such as shapes, radiance fields, or signed distance functions as continuous functions parameterized by neural networks, establishing an important foundation for generative modeling in continuous domains.} Furthermore, recent works have attempted to extend the framework of generative models  to function space, either by adapting existing algorithms to operate on discretized function data in $\mathbb{R}^n$ or by transforming input functions into a finite-dimensional latent space \cite{dutordoir2023neural, dupont2022data, phillips2022spectral, hui2022neural, bautista2022gaudi, chou2023diffusion}. 
Dupont et al. \cite{dupont2021generative} pioneered this direction by treating data points as continuous functions and learning distributions over them using adversarial training. Multiple theoretical frameworks have emerged for function diffusion models, including Gaussian measures on Hilbert spaces \cite{kerrigan2022diffusion}, dimension-independent bounds for infinite-dimensional models \cite{pidstrigach2023infinite}, generalized diffusion processes in function spaces \cite{franzese2024continuous}, and denoising diffusion operators with rigorous Gaussian processes and Langevin dynamics \cite{lim2023score}.Although these studies have demonstrated promising results for fixed discretizations, they generally struggle to scale to higher resolutions or to handle variable discretizations consistently,  as those encountered in scientific computing, partial differential equation (PDE) modeling, or operator learning~\cite{lu2021learning,kovachki2023neural,jiang2024fourier,yin2024scalable,lee2024efficient}.

% Although these approaches have shown promise for fixed discretizations, they face challenges in scaling to higher resolutions and handling variable discretizations. 
% Moreover, they lack a principled foundation for extending generative modeling to function spaces, where each sample represents a continuous object rather than a finite-dimensional vector. As a result, these approaches do not scale naturally to problems where the underlying data lies in an infinite-dimensional space, such as those encountered in scientific computing, partial differential equation (PDE) modeling, or operator learning~\cite{lu2021learning,kovachki2023neural,jiang2024fourier,yin2024scalable,lee2024efficient}.

{\color{black} In parallel, diffusion models have emerged as powerful tools for tackling inverse problems and modeling dynamical systems governed by PDEs. Recent works have developed methods for reconstructing spatiotemporal fields from sparse observations and partial supervision \cite{li2024learning, huang2024diffusionpde,shysheya2024conditional}. To address resolution limitations, several approaches have integrated neural operators with diffusion models \cite{rahman2022generative, oommen2024integrating, hu2024wavelet}, demonstrating improved performance in turbulence modeling and capture of multiscale phenomena. Furthermore, various strategies have attempted to incorporate physical priors directly into diffusion processes \cite{shu2023physics, qiu2024pi, bastek2024physics, gao2024bayesian, huang2024diffusionpde, zhuang2025spatially, jacobsen2025cocogen}. For instance, DiffusionPDE \cite{huang2024diffusionpde} introduces physics-based guidance during the posterior sampling stage, whereas PIDM \cite{bastek2024physics} embeds PDE residuals directly into the training objective. However, enforcing PDE residuals is highly sensitive to solution accuracy—small numerical errors can destabilize both training and inference. Moreover, physical constraints are inherently defined in continuous space, but these diffusion models operate on discretized representations. This mismatch introduces discretization artifacts, often leading to inaccurate or ill-posed PDE residuals, especially in inverse problems where key state variables are only partially observed.
These challenges underscore a fundamental gap between current data-driven generative models and the objective of producing physically consistent, continuous functions.}

To address these challenges, we propose FunDiff, a novel framework for physics-informed generative modeling over function spaces. Our contributions are as follows. {\color{black} We introduce a unified architecture that combines latent diffusion processes with function autoencoders, enabling the high-quality data generation
in a function-space representation.} This design supports evaluation at arbitrary spatial locations while seamlessly incorporating physical priors through the function space representation. We establish theoretical guarantees for our approach by proving that diffusion-based models attain minimax-optimal rates for density estimation in infinite-dimensional spaces under mild smoothness assumptions. We demonstrate the practical effectiveness of FunDiff across diverse applications in fluid dynamics and solid mechanics. Our experiments show that the method generates physically consistent samples with high fidelity to target distributions while maintaining robustness to noisy and low-resolution input data.
Taken together, our framework bridges the gap between theoretical guarantees in function spaces and practical neural network algorithms, establishing a principled approach to physics-informed generative modeling.

\section*{Results}

In this section,  we present theoretical results on generative models in function spaces, which establish the mathematical foundation of our work. Building on this theory, we introduce FunDiff, a practical framework for generating functions while incorporating physical priors.  
Then we demonstrate the effectiveness of our proposed FunDiff framework through a series of experiments on diverse physical systems. Specifically, we present four case studies with increasing complexity: (1) the Burgers' equation, showcasing the incorporation of PDE constraints; (2) Kolmogorov flow, highlighting divergence-free field generation; (3) linear elasticity, illustrating symmetry preservation; and (4) turbulent mass transfer, demonstrating conditional generation capabilities. For notations, interested readers are encouraged to refer to the Appendix \ref{app:notations}. {\color{black} 
It is important to emphasize that in all examples, FunDiff does not act as a neural PDE solver; rather, it learns to generate physically consistent continuous functions that approximate the distributions of solutions to selected PDE-governed systems when trained on such data. 
}

\subsection*{Generative models over function spaces}
\label{sec: theory}

In this section, we introduce a framework for generative modeling in function spaces. Given the inherently challenges of handling directly with infinitely-dimensional spaces, we begin by reducing the problem to a finite-dimensional setting, with preliminaries in Appendix \ref{app: Preliminaries}. By embedding functions into a finite-dimensional vector space, the task of generating new functions can be reformulated as 
a well-studied vector generation task.

To formalize this idea, let us begin with a vector space generative oracle $\mathcal{O}(d,n)$, which, given $n$ i.i.d. samples $\mathbf{x}^{1}, \mathbf{x}^{2}, \ldots, \mathbf{x}^{n}\in \mathbb{R}^d$ from an underlying distribution $\mu \in \mathcal{P}(\mathbb{R}^d)$, produces an estimate $\widehat{\mu}\in \mathcal{P}(\mathbb{R}^d)$ of $\mu$. This oracle is assumed to achieve the minimax statistical estimation rate
\[\bE\left[W_1(\widehat{\mu}, p)\right] \lesssim \varepsilon(d, n),\]
where the expectation is taken over the randomness of the samples, and $\varepsilon(d,n)$ aligns with the known minimax estimation rates \cite{weed2019sharp, dou2024optimal}. For the continuous density class $\mathcal C(\R^d) := \{\mu:\R^d\rightarrow [0,\infty)~:~ \mu~\text{is continuous}\}$, we have a minimax density estimation rate
\[\sup_{\widehat{\mu}}\inf_{\mu\in \mathcal C(\R^d)} \bE\left[W_1(\widehat{\mu}, \mu)\right] \asymp n^{-1/d}\]
for $d > 2$, where $\widehat{\mu}$ is generated by diffusion models in this paper. It is a special case for the $n^{-\frac{\alpha+1}{2\alpha+d}}$ minimax rate when the smoothness $\alpha=0$. Therefore, for the entire continuous density class $\mathcal C(\R^d)$, we have $\varepsilon(d, n) = n^{-1/d}$. 

Building upon this oracle, we seek to develop a generative modeling method that operates over a space of functions. Suppose that we have $n$ functions $f^{(1)}, f^{(2)}, \ldots, f^{(n)}:\mathbb{R}^d \to \mathbb{R}$, drawn i.i.d. from an unknown distribution $P \in \mathcal{P}(\mathcal{F})$, where $\mathcal{F}$ is a subset of the continuous function space $\mathcal{C}(\mathbb{R}^d)$. Our goal is to devise a theoretically sound methodology for generating new function samples from $P$, given only these observed samples.

To achieve this, we introduce a function autoencoder (FAE) framework. We define an encoder $\mathcal{E}:\mathcal{F}\to\mathbb{R}^D$, which maps a function to a finite-dimensional vector representation, and a decoder $\mathcal{D}:\mathbb{R}^D \to \mathcal{F}$, which maps these vectors back to functions. Applying $\mathcal{E}$ to the given functions $f^{(1)}, f^{(2)}, \ldots, f^{(n)}$ yields vector samples $\theta^{(1)}, \theta^{(2)}, \ldots, \theta^{(n)} \in \mathbb{R}^D$. We then use the vector-space generative oracle $\mathcal{O}(D,n)$ on these vectors to obtain an estimated distribution $\hat{p} \in \mathcal{P}(\mathbb{R}^D)$. Finally, we can sample new vectors $\theta \in \mathbb{R}^D$ from $\hat{p}$ and apply $\mathcal{D}$ to generate new functions $f = \mathcal{D}(\theta) \in \mathcal{F}$, the distribution of these newly generated functions is indicated by $\widehat{P}$ and serves as an estimation of $P$. This construction effectively leverages the theoretical foundations of vector-space generative modeling and extends them to the realm of infinite-dimensional function spaces.

In order to measure the distribution loss, we apply an analogy of the Wasserstein distance $W_1(P, \widehat{P})$. Notice that for distributions over the function space, the Wasserstein distance is defined similarly as
\[W_1(P, \widehat{P}) := \sup_{\mathrm{Lip}(\mathcal L)\leqslant 1} \left[\mathbb{E}_{f\sim P} \mathcal L(f) - \mathbb{E}_{f\sim \widehat{P}} \mathcal L(f)\right], \]
where $\mathcal L$ is a functional and its Lipschitz constant means:
\[\mathrm{Lip}(\mathcal L) = \sup_{f,g} \frac{\mathcal L(f) - \mathcal L(g)}{\|f-g\|_2}. \]
In the next part, we propose an upper bound of $W_1(P,\widehat{P})$ and explain all its factors.

\subsection*{Density estimation loss decomposition}

Suppose the vector representations $\{\theta^{(k)} = \mathcal{E}(f^{(k)}) \mid k\in[n]\}$ are drawn from a distribution $p = E_{\sharp}P$, which stands for the pushforward of $P$ under the encoder $\mathcal{E}$. Applying the vector-space generative oracle $\mathcal O(D,n)$ to these samples yields an estimated distribution $\widehat{p}\in \mathcal P(\R^D)$. We then feed $\hat{p}$ into the decoder $\mathcal{D}$, producing $G_\sharp \hat{p}$, an estimate of the original distribution $P$ in the function space. For the estimation loss, we have the bound
\begin{equation}
\label{eqn: framework}
\begin{aligned}
W_1(P, \widehat{P}) \leqslant \bE_{f\sim P} \|f - \mathcal{D}\circ \mathcal{E}(f)\|_2 + \mathrm{Lip}(\mathcal{D}) \cdot W_1(p, \widehat{p}). 
\end{aligned}
\end{equation}
Here, the first term $\bE_{f\sim P} \|f - \mathcal{D}\circ \mathcal{E}(f)\|_2$ represents the generalization error for the autoencoder $\mathcal{D} \circ \mathcal{E}: \mathcal F \rightarrow \mathcal F$. The second term $\mathrm{Lip}(\mathcal{D})$ stands for the Lipschitz constant of the decoder $\mathcal{D}$, defined by the inequality
\[\|\mathcal{D}(\theta)-\mathcal{D}(\theta')\|_2\leqslant \mathrm{Lip}(\mathcal{D})\cdot \|\theta-\theta'\|_2.\]
This quantity measures the robustness of $\mathcal{D}$ with respect to perturbations in the input vector space. The third term, $W_1(p, \widehat{p})$, reflects the accuracy of the density estimation for vector space, highlighting the performance of the oracle $\mathcal O(D,n)$ in estimating the true distribution $p$. The dimension of the feature space \(D\) must be chosen judiciously to strike a balance between ensuring accurate autoencoder reconstruction and mitigating the effects of the curse of dimensionality for the vector-space generative oracle. The proof of ~\eqref{eqn: framework} can be seen in Appendix \ref{sec:B.1}. Next, we propose some simple examples to help us understand the theoretical framework.

\paragraph{Illustrative example 1: Reproducing kernel Hilbert space (RKHS).} 
For a reproducing kernel Hilbert space (RKHS) $\mH$, we denote its eigenvalues as $\mu_1 \geqslant \mu_2 \geqslant \ldots$, and their corresponding eigenfunctions as $\psi_1, \psi_2, \ldots$. For any function $f\in \mH$, its eigen-decomposition has the form $f = \sum_{i=1}^{\infty} \theta_i \sqrt{\mu_i} \psi_i$ and its RKHS norm is defined as
$\|f\|_\mH^2 = \sum_{i=1}^{\infty} \theta_i^2$. Define the RKHS unit ball as $\mathcal B := \{f\in\mH:\|f\|_\mH \leqslant 1\}$. 

Given $n$ sample functions $f^{(1)}, f^{(2)}, \ldots, f^{(n)}\sim P \in \mathcal P(\mB)$, assume their eigen-decomposition is
\[f^{(k)} = \sum_{i=1}^{\infty} \theta_i^{(k)}\cdot \sqrt{\mu_i}\psi_i, ~~~(k=1,2,\ldots, n).\]
We define the encoder $\mathcal{E}: \mB \rightarrow \R^D$ as the truncation of $D$ coefficients and the decoder $\mathcal{D}: \R^D\rightarrow \mB$ as the linear combination:
\[\mathcal{E}: f \mapsto (\theta_1, \theta_2, \ldots, \theta_D)^\top, ~~~\mathcal{D}: (\theta_1, \theta_2, \ldots, \theta_D)^\top \mapsto \sum_{i=1}^D \theta_i\cdot \sqrt{\mu_i}\psi_i. \]
Hence, the autoencoder $\mathcal{D}\circ \mathcal{E}$ effectively serves as a projection (truncation) onto the subspace spanned by the first $D$ eigenfunctions. Next, in order to upper bound the density estimation loss $W_1(P, \widehat{P})$, we introduce two different assumptions corresponding to the polynomial and exponential eigenvalue decay of the RKHS $\mH$. 

\begin{assumption}[$2\beta$-polynomial decay]
Assume that the RKHS $\mH$ is embedded to a $\beta$-th order Sobolev space where $\beta>\frac12$. It leads to the property that the eigenvalues with respect to the kernel function $K(\cdot, \cdot)$ have the property of $2\beta$-polynomial decay as $\mu_i \asymp i^{-2\beta}$ for all $i\geqslant 1$. To be more specific, we assume that
\[c_{\mH}\cdot i^{-2\beta} \leqslant \mu_i \leqslant C_{\mH}\cdot i^{-2\beta}\]
holds for $\forall i\geqslant 1$, where $c_{\mH}$ and $C_{\mH}$ are constants.
\label{assump-2}
\end{assumption}

\begin{assumption}[$\gamma$-exponential decay]
Assume that the eigenvalues with respect to the kernel function $K(\cdot, \cdot)$ have the property of $\gamma$-exponential decay as $\mu_i \asymp \cdot\exp(-C_1\cdot i^{\gamma})$ for all $i\geqslant 1$ where $C_1>0$ is a positive constant. To be more specific, we assume that
\[c_{\mH}\cdot \exp(-C_1\cdot i^{\gamma}) \leqslant \mu_i \leqslant C_{\mH}\cdot \exp(-C_1\cdot i^{\gamma})\]
holds for $\forall i\geqslant 1$, where $c_{\mH}$ and $C_{\mH}$ are constants. 
\label{assump-2-exp}
\end{assumption}

Then, the reconstruction loss $\E_{f\sim P} \|f-\mathcal{D}\circ \mathcal{E}(f)\|_2$ represents the truncation loss under the RKHS eigen-decomposition, and the robustness $\mathrm{Lip}(\mathcal{D})$ is directly related to the RKHS norm bound of $\mathcal B$.

\begin{theorem}[Density estimation over RKHS space]
\label{thm: 1}
Assume $P$ is a distribution supported on the RKHS unit ball $\mathcal B := \{f\in\mH:\|f\|_\mH \leqslant 1\}$. By using the generative oracle $\mathcal O(D, n)$ and the function autoencoder (FAE) framework introduced above, the density estimation loss over RKHS space can be bounded as follows: \\
Under the polynomial decay setting:
\[W_1(P, \widehat{P}) \lesssim \left(\frac{\log\log n}{\log n}\right)^{\beta}.\]
Under the exponential decay setting:
\[W_1(P, \widehat{P}) \lesssim \exp\left(-(C_1/2)^{\frac{1}{1+\gamma}}(\log n)^{\frac{\gamma}{1+\gamma}}\right),\]
after we choose an optimal latent dimension $D$. 
\end{theorem}
\begin{proof}
The proof is provided in Appendix \ref{sec:proof.1}. 
\end{proof}

\paragraph{Illustrative example 2: Barron space.}

First, we introduce the concept of Barron space \cite{barron1993universal}. For function $f: \Omega:=[0,1]^d\rightarrow\bR$, we aim to recover $f$ using a fully connected two-layer neural network with ReLU activation:
\[f(\bx; \theta)=\sum_{k=1}^{m}a_{k}\sigma(\tb_{k}\cdot\bx+c_{k}).\]
Here, $\sigma(\cdot)$ is the component-wise ReLU activation $\sigma(t)=\max(0,t)$, $\tb_{k}\in \bR^{d}, a_k, c_k \in \bR$ for $k\in [m]$, and the whole parameter set $\theta = \{(a_{k},\tb_{k},c_{k})\}_{k=1}^{m}$ is to be learned, and $m$ is the width of the network. As we can see, $\theta\in \bR^{m(2+d)}$. In order to control the magnitude of the learned network, we use the following scale-invariant norm.
\begin{definition}[Path norm \cite{neyshabur2015norm}]
For a two-layer ReLU network, the path norm is defined as
\[\|\theta\|_{\mathrm P} = \sum_{k=1}^{m}|a_{k}|\cdot (\|\tb_{k}\|_{1}+|c_{k}|). \]
\end{definition}
\begin{definition}[Spectral norm]
Given $f\in L^{2}(\Omega)$, denote by $F\in L^{2}(\R^d)$ an extension of $f$ to $\R^d$. Let $\hat{F}$ be the Fourier transform of $F$, then
\[f(\bx) = \int_{\bR^{d}}e^{i\langle\bx,\omega\rangle}\hat{F}(\omega)\,\rd\omega, ~~\forall\bx\in\Omega.\]
We define the spectral norm of $f$ by
\[\gamma(f)=\inf\limits_{F\in L^{2}(\bR^d),F|_{\Omega}=f|_{\Omega}}\int_{\R^d}\|\omega\|_{1}^{2}\cdot |\hat{F}(\omega)|\,\rd\omega.\]
\end{definition}
\begin{assumption}
We consider all functions with a bounded value and a bounded spectral norm.
\[\mathcal F_{s}=L^{2}(\Omega)\cap\{f(\bx):\Omega\rightarrow\bR\mid\gamma(f)\leqslant 1, \|f\|_{\infty}\leqslant 1\}.\]
We let the function class $\mathcal F = \mathcal F_{s}$, which makes the distribution $P$ supported on Barron space $\mathcal F_{s}$. 
\end{assumption}
In the following analysis, we mainly rely on the approximation property for two-layer ReLU networks over the Barron space $\mathcal F_s$, which are proposed in Refs.~\cite{barron1993universal,breiman1993hinging,klusowski2016risk}.
\begin{lemma}
\label{lemma: approx-Barron}
For any $f\in \mathcal F_{s}$, there exists a two-layer network $f(\bx; \tilde{\theta})$ of width $m$ such that:
\[\left\|f(\bx; \tilde{\theta}) - f(\bx)\right\|_{\infty}^{2}\leqslant \frac{16\gamma^{2}(f)}{m}\leqslant \frac{16}{m}. \]
Furthermore, the parameter $\tilde{\theta}$ can be bounded as:
\[\|\tb_k\|_1 = 1, |c_k|\leqslant 1,~~\text{and}~~|a_k|\leqslant \frac{2\gamma(f)}{m}\leqslant \frac2m~~~~\text{holds for~~}\forall k\in [m]. \]
\end{lemma}
In the context of the FAE framework, we set our encoder as $\mathcal{E}: f \mapsto \tilde{\theta}$. If there exist multiple valid $\tilde{\theta}$ for $f\in F_s$, we can choose any one of them as the encoder's output. As we see, the encoder $\mathcal{E}$ maps the function of the Barron space to a parameter set with bounded path norm. The decoder $\mathcal{D}: \theta\mapsto f(\bx; \theta)$ is simply a parameter injection of a two-layer ReLU network. Hence, the autoencoder $\mathcal{D}\circ \mathcal{E}$ effectively serves as a two-layer ReLU network approximation. Next, we upper bound the density estimation loss $W_1(P, \widehat{P})$.

\begin{theorem}[Density estimation over Barron space] 
\label{thm: 2}
Assume $P$ is a distribution supported on the Barron space $\mathcal F_s$. By using the generative oracle $\mathcal O(D, n)$ and the function autoencoder (FAE) framework introduced above, the density estimation loss over Barron space can be bounded as
\[W_1(P, \widehat{P}) \lesssim \sqrt{\frac{d \log d}{\log n}}.\]
\end{theorem}
\begin{proof}
The proof is provided in Appendix \ref{sec:proof.2}. 
\end{proof}
{
\color{black}
Now we have proved the statistical upper bound of density estimation over the RKHS ball and Barron space. To further show the optimality of such an upper bound, we provide a statistical lower bound of density estimation over the RKHS ball, which matches its upper bound with only a negligible $\mathcal O(\mathrm{poly}(\log\log n))$ gap.
\begin{theorem}[Minimax lower bound for Wasserstein estimation over an RKHS ball]
\label{thm: lower}
Assume $\mathcal H$ is an RKHS following $2\beta$-polynomial eigenvalue decay (Assumption \ref{assump-2})
\[c_{\mH}\cdot i^{-2\beta} \leqslant \mu_i \leqslant C_{\mH}\cdot i^{-2\beta}\]
and define the RKHS unit ball $\mathcal B := \{f\in \mathcal H~:~\|f\|_{\mH}\leqslant 1\}$. 
We observe $n$ i.i.d. samples $X_1, X_2, \ldots, X_n \sim P$, where the unknown distribution $P$ is supported on the RKHS ball $\mathcal B$. Let the statistical minimax rate
\[\mathcal R_n := \inf_{\widehat{P}_n} \sup_{P:~\mathrm{supp}(P)\subseteq \mathcal B} \bE \left[W_1(P, \widehat{P}_n)\right]. \]
Then there exists a constant $K(\beta, c, C) > 0$, depending only on pure constants $\beta, c, C$ in the eigenvalue decay assumption, such that for all sufficiently large $n$, 
\[\mathcal R_n \geqslant K(\beta, c, C)\cdot \left(\frac{1}{\log n}\right)^\beta. \]
\end{theorem}
\begin{proof}
The proof is provided in Appendix \ref{app:lower}.
\end{proof}
Notice that its corresponding statistical upper rate is $\left(\frac{\log\log n}{\log n}\right)^\beta$ according to our results in Theorem \ref{thm: 1}. It means that our upper and lower bounds only have a $\mathcal O(\mathrm{poly}(\log\log n))$ gap, which is negligible to our statistical rate. 
}

\subsection*{FunDiff in practice} 
\begin{figure*}
    \centering
    \includegraphics[width=1.0\linewidth]{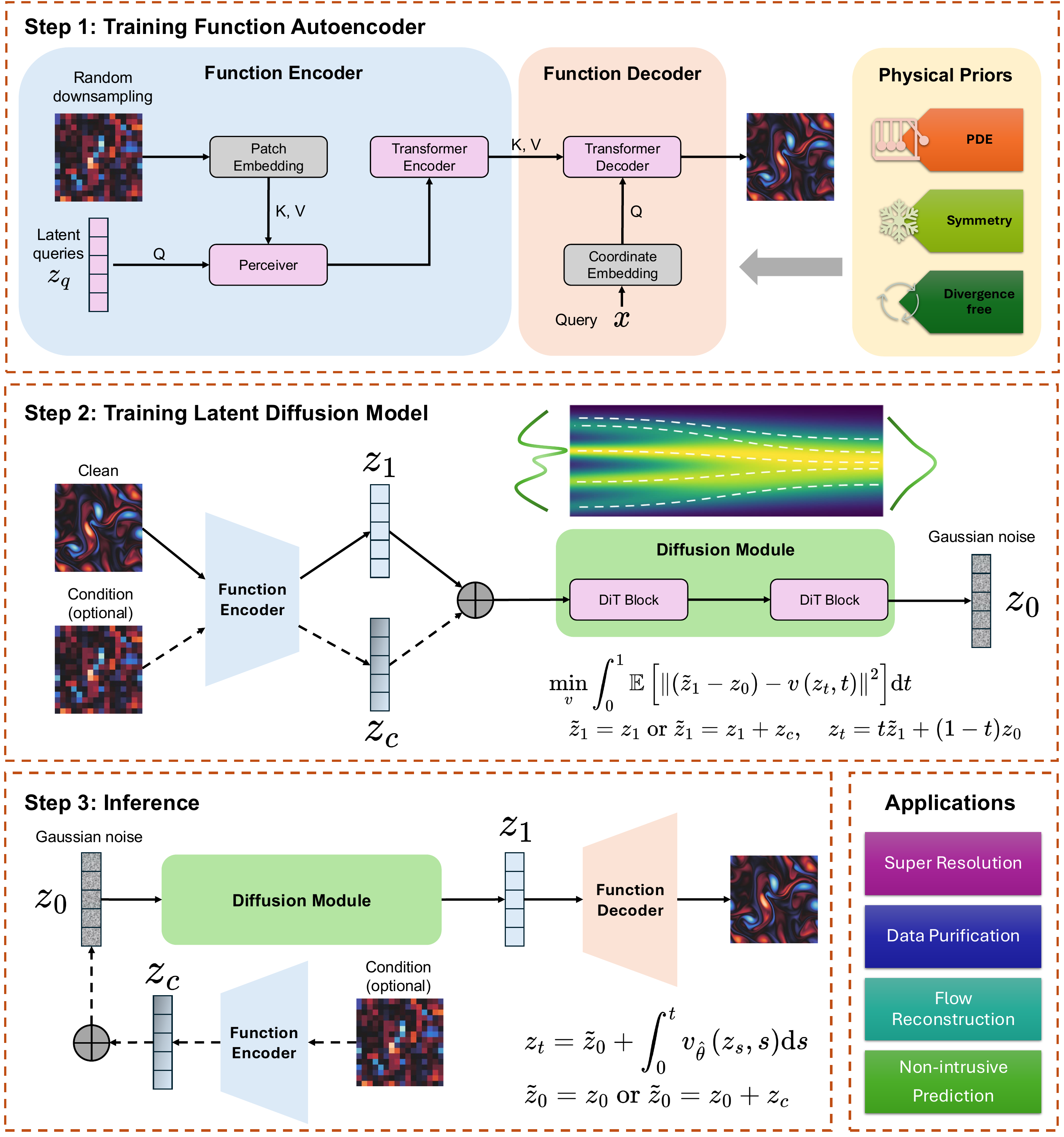}
    \caption{\textbf{Overview of the function generative framework.} The framework comprises three main steps. (1) We first train a function autoencoder (FAE) that maps discretized function data into a continuous latent space while preserving physical constraints. The encoder employs a Perceiver module to handle variable discretizations and project them into a unified latent representation. A Vision Transformer (ViT) processes the discretized inputs, and the decoder enables continuous function evaluation via cross-attention between query coordinates $x$ and encoded features. When known physical priors are available, the decoder is modified accordingly. (2) Next, we train a Diffusion Transformer (DiT) in the learned latent space to generate diverse function samples. For conditional generation, partial observations are encoded using the pretrained transformer and incorporated into the diffusion model inputs. The diffusion model is trained using rectified flow. (3) During inference, latent codes are sampled by solving a reverse-time ODE from Gaussian noise, and then decoded into continuous functions via the decoder. The framework naturally supports various physical constraints through its continuous latent representation, enabling applications in super-resolution, flow reconstruction, data purification, and non-intrusive prediction.}
    \label{fig:pipeline}
\end{figure*}

To realize the framework of function generative models outlined above, we propose FunDiff (Fig.~\ref{fig:pipeline}), which consists of a novel FAE and a latent diffusion model to generate continuous functions. The FAE learns the mapping from discrete function representations to continuous ones, while also enabling the enforcement of prior physical constraints.
The latent diffusion model is then trained directly on the latent manifold learned by the pretrained FAE encoder.
Our generation process follows a simple two-step procedure: we first sample latent codes by integrating the reverse-time ODE from Gaussian noise to the target data distribution, and then transform these codes into continuous functions via the FAE decoder.

\paragraph{Physical prior integration.}
The FAE framework enables flexible integration of physical priors through multiple mechanisms of either soft or hard constraints. For general PDE constraints, we augment the loss with a residual term:
\begin{equation*}
\mathcal{L}_{\text{PDE}} = |\mathcal{R}(\hat{f})|_2^2,
\end{equation*}
where $\mathcal{R}$ represents the PDE residual and $\hat{f}$ is the reconstructed function. 
The continuous representation of FAE enables automatic differentiation to compute all required derivatives, allowing constraint enforcement across the entire domain.   {\color{black} It is worth noting that these physics-informed terms serve as regularizers rather than numerical solvers of the governing equations; they guide the learned FAE to produce output functions that are physically consistent.}

Furthermore, some physical properties can be incorporated through architectural modifications.
For periodic boundary conditions, one approach is to modify the coordinate embedding with Fourier bases of specific frequencies~\cite{lu2021physics}.
When handling divergence-free fields, particularly important in fluid dynamics applications, the FAE can be configured to learn a stream function $\psi$, from which velocity field components are derived as $u=\frac{\partial \psi}{\partial y}, \quad v=-\frac{\partial \psi}{\partial x}$, thereby ensuring divergence-free flow. Symmetry constraints can be incorporated when needed through modifications to the decoder's final layer.

\paragraph{Related work.} We emphasize that our approach fundamentally differs from existing physics-informed diffusion models \cite{shu2023physics, qiu2024pi, bastek2024physics, gao2024bayesian, huang2024diffusionpde, zhuang2025spatially, jacobsen2025cocogen}, which impose physical constraints either by modifying the diffusion model's training objective or by guiding the posterior sampling process. These methods operate on discretized outputs, where physical laws—typically enforced through PDE residuals—are subject to discretization errors. This not only leads to inaccurate constraint enforcement, but also introduces additional optimization challenges. Moreover, in inverse problems where key physical variables are unobserved, such residuals may be undefined at inference time, further limiting the applicability of these approaches.
In contrast, our method enforces physical constraints within the FAE, which is fully decoupled from the training and inference processes of the latent diffusion model. Since the FAE provides a continuous neural representation, we can compute PDE residuals by automatic differentiation when needed. This design enables both exact and soft enforcement of constraints in continuous space, mitigating discretization-induced errors, and enabling more stable and physically consistent generation.
Details on how to impose these constraints are provided in the following examples through comprehensive numerical experiments, demonstrating that our framework can generate diverse function samples while naturally preserving the underlying physical principles.

\subsection*{Learning 1D damped sinusoidal functions}

\begin{figure*}
    \centering
    \includegraphics[width=1.0\linewidth]{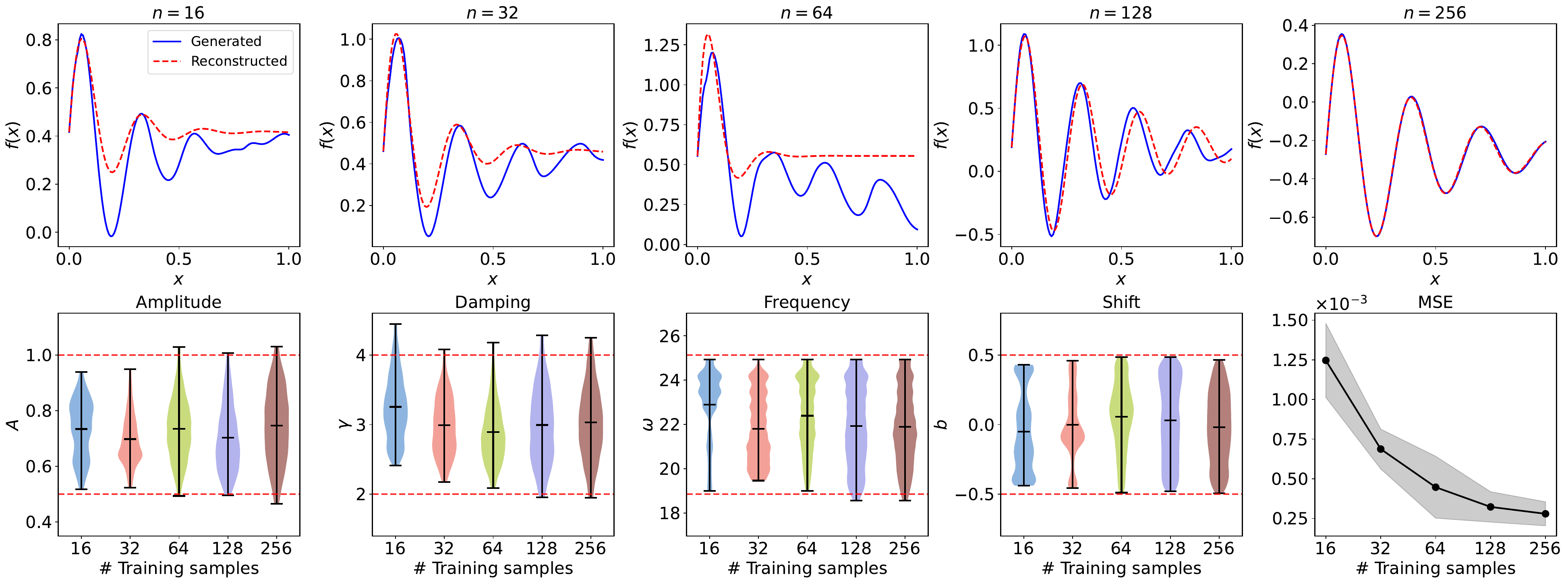}
    \caption{\textbf{Damped sinusoidal functions.} Analysis of generative quality across models trained with different numbers of samples ($n=16,32,64,128,256$).
    (\textbf{Top}) Comparison between the worst-case generated samples (blue dots) and their reconstructed signals (red dashed lines) obtained through parameter estimation. (\textbf{Bottom}) The first four panels show the estimated parameter distributions for amplitude ($A$), damping coefficient ($\gamma$), angular frequency ($\omega$), and vertical shift ($b$). Red dashed lines indicate the true parameter bounds from the underlying uniform distributions. The rightmost panel displays the MSE of the reconstructed signal. The results demonstrate that models trained with more samples generate functions that better align with the true distributions and achieve lower reconstruction error.}
    \label{fig:damp}
\end{figure*}

To validate our theoretical analysis and demonstrate the training pipeline of the proposed framework, we first present a pedagogical example using a simple distribution of one-dimensional damped sinusoidal signals. The target functions are defined as
\begin{align}
\label{eq:damp}
f(x) = A \exp(-\gamma x) \sin(\omega x) + b, \quad x \in [0, 1],
\end{align}
where the parameters are sampled from uniform distributions: amplitude $A \sim \mathcal{U}[0.5, 1]$, decay rate $\gamma \sim \mathcal{U}[2, 4]$, angular frequency $\omega \sim \mathcal{U}[6\pi, 8\pi]$, and vertical offset $b \sim \mathcal{U}[-0.5, 0.5]$.

We employ an FAE with a 6-layer encoder and a 2-layer decoder, each with a hidden dimension of 256 and 8 attention heads. The FAEs are trained on datasets of varying sizes, with function samples discretized at 128 uniform grid points. We train the FAE by minimizing the loss using the Adam optimizer for $5 \times 10^4$ iterations. The learning rate follows a warmup and exponential decay schedule: it linearly increases from 0 to $10^{-3}$ over the first 1,000 steps, then exponentially decays every 1,000 steps with a decay rate of 0.9. Following the FAE training, we train the latent diffusion model by minimizing ~\eqref{eq:diffusion_loss} using an identical learning rate schedule.

For evaluation, we generate 8,192 sample functions using our inference procedure with 100 denoising steps. We assess the quality of these generated samples through parameter estimation, leveraging the analytical form of damped sinusoidal functions. Specifically, for each generated sample, we perform a least-squares fitting using ~\eqref{eq:damp} to obtain the inferred parameters and reconstruct the corresponding signal. This approach enables a rigorous quantitative assessment of our model's ability to capture the underlying function space through two metrics: the comparison between reconstructed and generated signals, and the analysis of inferred parameter distributions against the target uniform distributions.

We provide a comprehensive evaluation of our model's generative performance across varying training dataset sizes (Fig.~\ref{fig:damp}). The top row illustrates the worst-case generated samples alongside their reconstructions via parameter estimation. As the number of training samples increases, the quality of generation improves significantly. The bottom row analyzes the parameter distributions, revealing the evolution of our model's capacity to capture the underlying function space.  With limited training data (16 or 32 samples), the estimated parameter distributions exhibit narrower ranges or deviate significantly from the target uniform distributions, indicating insufficient coverage of the parameter space. 
As the size of training dataset increases, the estimated parameter distributions progressively converge to the target uniform distributions, with the model trained on 256 samples achieving notably better alignment with the true distributions. 
This improvement is quantitatively validated by the decreasing mean squared error (MSE) between the generated samples and their reconstructions. These empirical results strongly support our theoretical analysis.

\subsection*{Kolmogorov flow}

In this example, we use the proposed FunDiff to generate divergence-free velocity fields, a fundamental requirement for ensuring the physical consistency of incompressible flow simulations in computational fluid dynamics. We consider two scenarios: unconditional flow generation and flow reconstruction from partial observations. To evaluate our approach, we use the two-dimensional Kolmogorov flow, a well-established benchmark for analyzing transition to turbulence and the stability of incompressible fluid systems. The governing equations are formulated in the vorticity-velocity representation as follows:
\begin{align*}
    \frac{\partial \omega(x, t)}{\partial t} 
    + \mathbf{u}(x, t) \cdot \nabla \omega(x, t) 
    &= \frac{1}{\operatorname{Re}} \nabla^2 \omega(x, t) + f(x), \\
    \nabla \cdot \mathbf{u}(x, t) 
    &= 0, \\
    \omega(x, 0) 
    &= \omega_0(x),
\end{align*}
where $\omega$ is the vorticity, $\mathbf{u} = (u, v)$ is the velocity field, and $\operatorname{Re}$ is the Reynolds number, set to 1000. The forcing term $f(x)$ drives the flow and $x = [x_1, x_2]$. The equations are defined with periodic boundary conditions in the spatial domain $x \in (0, 2\pi)^2$ and time interval $t \in (0, T]$.
The forcing term is defined as
\begin{align*}
    f(x)=-4 \cos \left(4 x_2\right)-0.1 \omega(x, t),
\end{align*}
where the additional drag force term $0.1 \omega(x, t)$ prevents energy accumulation on large scales.

To strictly impose periodic boundary conditions and ensure the divergence-free condition, we modify the forward pass of the FAE decoder as follows:
\begin{align}
    \left[x, y, \mathcal{E}_\theta(f)\right] \xrightarrow{\mathcal{D}_\theta}\psi(\gamma(x, y)),
\end{align}
where $\mathcal{D}_\theta$ denotes the forward pass of the decoder, and the coordinate embedding $\gamma$ is defined as
\begin{align}
    \gamma(x, y)=[\cos (x), \sin (x), \cos(y), \sin(y)].
\end{align}
This embedding guarantees that the output functions have a periodicity of $2 \pi$ within $[0,2 \pi]^2$. An incompressible velocity field is then computed as $(u, v)=(\partial \psi / \partial y,-\partial \psi / \partial x)$.

\subsubsection*{Unconditional flow generation}

\begin{figure*}
    \centering
    \includegraphics[width=0.9
    \linewidth]{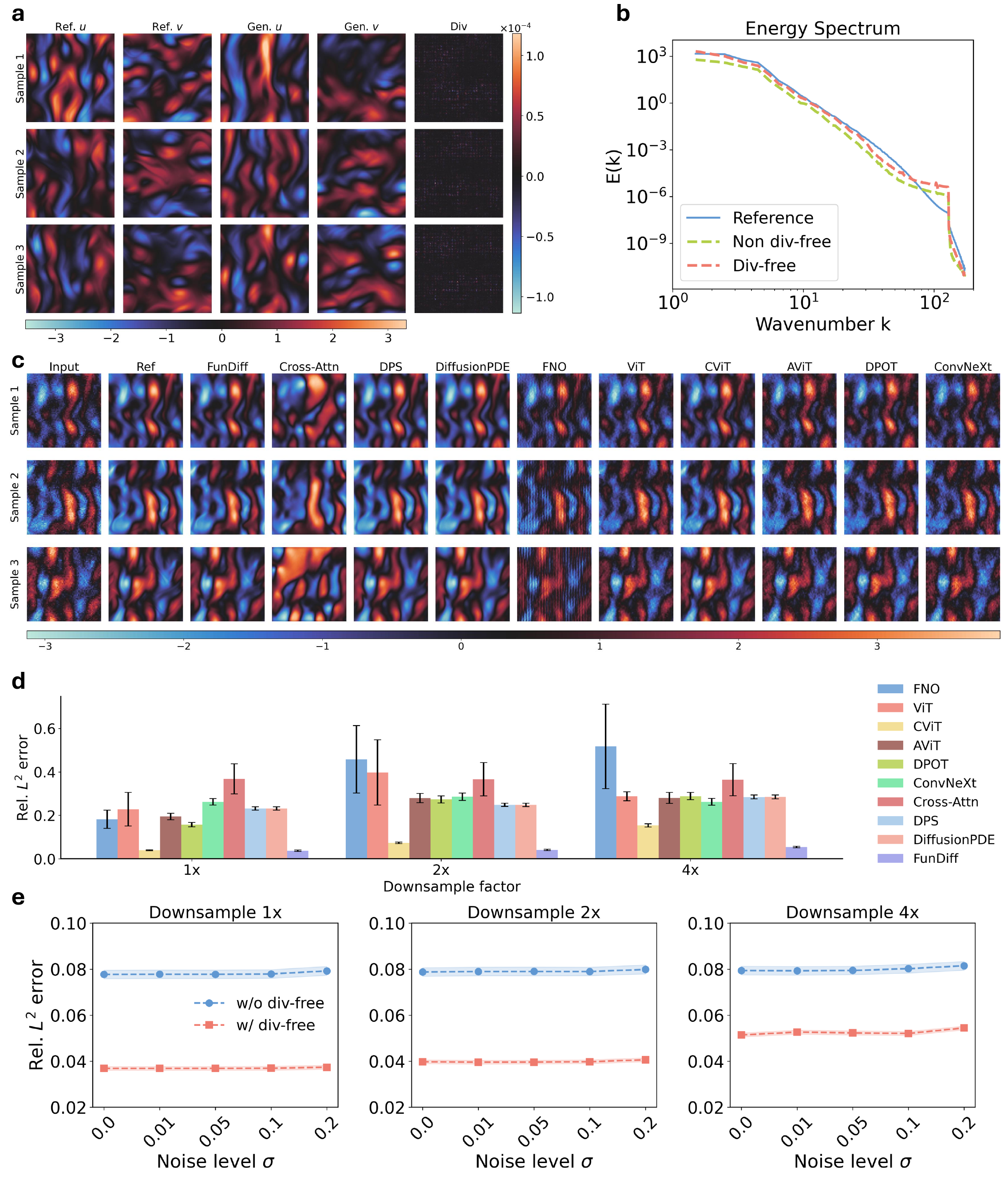}
    \caption{\textbf{Kolmogorov flow.} (\textbf{a}) Unconditional generation with enforced divergence-free constraint. (\textbf{b}) Comparison of spectral energy distributions between models constrained with and without divergence-free conditions. \textcolor{blue}{(\textbf{c}) Flow field reconstruction from noisy, low-resolution observations. Visualization of reconstructed velocity fields using different methods with input data of 64 × 64 resolution corrupted by 20\% Gaussian noise. (\textbf{d}) Comparative analysis of reconstruction errors across methods at various downsampling factors.} (\textbf{e}) Relative $L^2$ error of reconstructed velocity fields across varying noise levels ($\sigma$) for different spatial resolution reductions (1x, 2x, and 4x downsampling). Dashed lines show mean performance with (blue) and without (red) PDE constraints, while shaded regions indicate one standard deviation.}
    \label{fig:kf_combined}
\end{figure*}

Following the training procedure in the Method section, we present our results in Fig.~\ref{fig:kf_combined}a, which displays representative generated samples along with their corresponding divergence fields, computed via automatic differentiation. The predicted divergence values are negligible, confirming the effectiveness of our approach. We also compare the spectral energy distributions between the training dataset and 1,000 generated samples (Fig.~\ref{fig:kf_combined}b). The spectra show excellent agreement up to wavenumbers of $\mathcal{O}(10^2)$. However, the model leads to some artificial high-frequency components at larger wavenumbers.

To demonstrate the importance of the divergence-free constraint, we conduct a comparative study using an unconstrained FAE trained directly on the velocity field. In this case, the forward pass is formulated as
\begin{align*}
\left[x, y, \mathcal{E}_\theta(f)\right] \xrightarrow{\mathcal{D}_\theta}[u(x, y), v(x, y), \omega(x, y)].
\end{align*}
This unconstrained approach results in significantly higher divergence values (Fig.~\ref{fig:kf_non_div_samples}). Additionally, the spectral energy shows some discrepancies compared to the model constrained by the exact divergence-free condition.

\subsubsection*{Flow reconstruction from partial observations}

For our second case study, we evaluate the model's capability to reconstruct flow fields from corrupted partial observations. We use 90\% of the data for training and the rest for testing. To improve resolution-invariant reconstruction capabilities, we implement a dynamic downsampling strategy during the training of the FunDiff components. Specifically, at each training iteration, the input velocity field is randomly downsampled by factors of 1, 2, or 4. We first examine the impact of the divergence-free constraint by comparing FunDiff models with and without this condition (Fig.~\ref{fig:kf_combined}e). Both variants demonstrate robust performance across different downsampling factors and noise levels. Notably, the divergence-constrained FunDiff model consistently achieves lower test errors, validating the importance of physical constraints in the reconstruction process.

{\color{black}
Furthermore, we benchmark our approach against several state-of-the-art neural operators and PDE surrogate models, including FNO~\cite{li2021fourier}, ViT~\cite{dosovitskiy2020image}, and UNet~\cite{ronneberger2015u}, as well as more recent architectures such as CViT~\cite{wang2024cvit}, DPOT~\cite{hao2024dpot}, CNextU-Net~\cite{ohana2024well}, and AViT~\cite{mccabe2023multiple}. To further evaluate performance under inverse problem settings, we also include three diffusion-based baselines: Diffusion Posterior Sampling (DPS)~\cite{chung2023diffusion}, DiffusionPDE~\cite{huang2024diffusionpde}, and Spatially-Aware Diffusion~\cite{zhuang2025spatially}. }
Detailed implementation specifications for these baseline models are provided in Appendix \ref{app:baselines}. Fig.~\ref{fig:kf_combined}c presents reconstructed flow fields from various models, where the input consists of 64 × 64 resolution velocity fields corrupted by 20\% Gaussian noise. {\color{black}
It can be observed that neural operator baselines tend to produce highly discontinuous or spatially inconsistent patterns, whereas diffusion-based models recover coherent large-scale structures but often lose fine-scale features.
}
The results demonstrate that our model uniquely achieves both smooth and accurate reconstructions. Additional visualizations of our model's predictions, with and without the divergence-free condition imposed in FAE decoder, are shown in Figs.~\ref{fig:kf_samples} and~\ref{fig:kf_no_div_free_samples}, respectively. The quantitative comparison using relative $L^2$ errors across different downsampling factors is presented in Fig.~\ref{fig:kf_combined}e. These results confirm that our FunDiff model maintains consistent performance across all tested resolutions while exhibiting strong resilience to noise.

\subsection*{Burgers' equation}

As our next example, we demonstrate the effectiveness
and flexibility of our framework in generating physically consistent samples through the incorporation of physical laws as soft constraints in the training of FAE. To this end, we consider the Burgers' equation, which serves as a fundamental prototype for conservation laws that can develop discontinuities. The governing equation is given by
\begin{align}
    \label{eq:burgers}
    & \frac{d u}{d t}+u \frac{d u}{d x}-\nu \frac{d^2 u}{d x^2}=0, \quad(x, t) \in(0,1) \times(0,1], \\
& u(x, 0)=a(x), \quad x \in(0,1)
\end{align}
with periodic boundary conditions, where $\nu=0.001$ denotes the viscosity coefficient, and the initial condition $a(x)$ is sampled from a Gaussian random field (GRF) satisfying periodic boundary conditions.

Let $u_\theta$ denote the FAE output. We define the PDE residual as
\begin{align*}
    \mathcal{R}[u_\theta](x, t) = \frac{d u_\theta(x, t)}{d t} + u_\theta(x, t) \frac{d u_\theta(x, t)}{d x}-\nu \frac{d^2 u_\theta(x, t)}{d x^2}.
\end{align*}
The FAE is trained by minimizing a composite loss
\begin{align*}
\mathcal{L}(\mathbf{\theta}) := \underbrace{w_{data}\frac{1}{NM} \sum_{i=1}^{N} \sum_{j=1}^{M}  \left| {u}_{\mathbf{\theta}}^i(x_j, t_j) - {u}^i(x_j, t_j) 
 \right|^2}_{\mathcal{L}_{data}(\mathbf{\theta})}  + \underbrace{w_{physics}\frac{1}{NM} \sum_{i=1}^{N} \sum_{j=1}^{M} \left|  \mathcal{R}[{u}_{{\theta}}^i](x_j, t_j)  \right|^2}_{\mathcal{L}_{physics}(\mathbf{\theta})}. \
\end{align*}
Here, for each sample $u^i$, we randomly sample query coordinates $\{(x_j, t_j)\}_{j=1}^M$ from the uniform grid during each gradient descent iteration. All network derivatives are computed using automatic differentiation. Specifically, we set $w_{data}=100, w_{physics}=1$ and randomly downsample $u^{i}$ by a factor of 1, 2, 5 at each iteration of gradient descent.

\begin{figure*}
    \centering
    \includegraphics[width=0.95\linewidth]{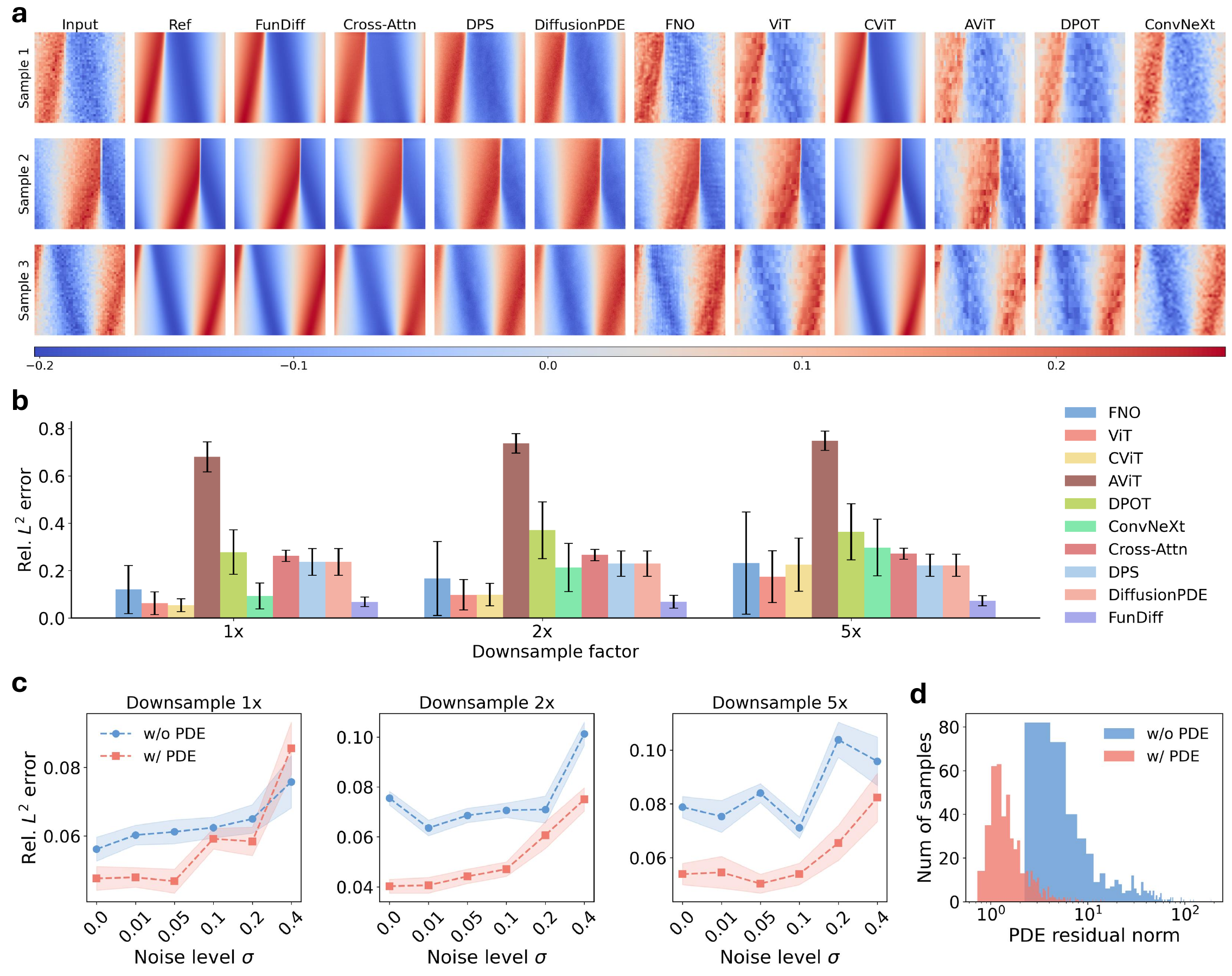}
    \caption{\textbf{Burgers' equation.} \textcolor{blue}{(\textbf{a})
    Solution reconstruction from noisy, low-resolution observations. Visualization of reconstructed solutions using different methods with input data of 40 × 40 resolution corrupted by 20\% Gaussian noise. (\textbf{b}) Comparative analysis of reconstruction errors across methods at various downsampling factors.} (\textbf{c})
    Relative $L^2$ error of reconstructed solutions across varying noise levels ($\sigma$) for different spatial resolution reductions (1x, 2x, and 5x downsampling). Dashed lines show mean performance with (blue) and without (red) PDE constraints, while shaded regions indicate one standard deviation. (\textbf{d}) Histogram of PDE residual norms, comparing the physical consistency of solutions with and without PDE constraints, demonstrating improved physical consistency when PDE constraints are incorporated.}
    \label{fig:burgers_combined}
\end{figure*}

{\color{black}
Similar to the Kolmogorov flow example, our objective here is to train a FunDiff model to reconstruct the solution field from sparse observations.
}
We evaluate our model against established baseline methods. We show three examples in Fig.~\ref{fig:burgers_combined}a. {\color{black} Neural operator baselines generally produce blurry and discontinuous reconstructions, whereas diffusion-based approaches yield smoother results but often exhibit small oscillations near sharp gradients.} In contrast, our FunDiff generates sharp and smooth reconstructions, which are in excellent agreement with the reference numerical solutions. Fig.~\ref{fig:burgers_combined}b presents a quantitative comparison of model performance across inputs with varying downsampling factors and 20\% Gaussian noise corruption. Notably, FunDiff outperforms all baselines.
Additional examples of our model’s predictions are shown in Fig.~\ref{fig:burger_pde_samples}. Their physical consistency is confirmed by low PDE residuals, remaining on the order of $\mathcal{O}(10^{-2})$ across the entire domain. We also systematically examine the effect of noise and data sparsity (Fig.~\ref{fig:burgers_combined}c). As expected, reconstruction error increases with higher noise levels, but FunDiff demonstrates strong robustness to data sparsity, maintaining stable performance even under aggressive downsampling (factors up to 5).

To rigorously evaluate the impact of physical constraints, we conduct an ablation study by training the function encoder without the physics loss term $\mathcal{L}_{physics}$. The absence of PDE-based regularization leads to significantly higher predictive errors and PDE residuals in the generated solutions (Fig.~\ref{fig:burgers_combined}d). This degradation in physical consistency is further visualized through the generated samples in Fig.~\ref{fig:burgers_no_pde_samples}, underscoring the crucial role of physics-informed training in our framework.

\subsection*{Linear elasticity}

Symmetries play an important role in physics, providing a framework for understanding conservation laws and invariances that govern natural phenomena. This example aims to demonstrate how to embed prior symmetries into FunDiff models.
To this end, we consider linear elasticity in computational micromechanics, which involves solving the Lippmann-Schwinger equation \cite{LSequation} in a 2D periodic representative volume element (RVE) microstructural domain $\Omega \subset \mathbb{R}^2$ with spatial coordinates $\mathbf{x}=(x, y)$ under plane strain conditions. The problem is formulated as
\begin{equation}
\begin{dcases*}
\varepsilon_{i j}(\mathbf{x})+\int_{\Omega} G_{i j k l}^{(0)}\left(\mathbf{x}, \mathbf{x}^{\prime}\right): [\sigma_{i j}\left(\mathbf{x}^{\prime}\right)-C_{i j k l}^{0}:\varepsilon_{k l}\left(\mathbf{x}^{\prime}\right)] \mathrm{d} \mathbf{x}^{\prime}-\bar{\varepsilon}_{i j}=0, \quad \mathbf{x} \in \Omega, \\
\sigma_{i j}(\mathbf{x})=C_{i j k l}(\mathbf{x}): \varepsilon_{k l}(\mathbf{x}). \\
\end{dcases*}
\label{setup:dotLS}
\end{equation}
These relations represent balance and constitutive equations, respectively, which are written in a component-wise manner ($ 1\leq i, j, k, l \leq 2$) using the Voigt notation. $\mathbf{x}$ and $\mathbf{x}^{\prime}$ denote a pair of target and source points inside the RVE domain $\Omega$. $\sigma_{i j}, \varepsilon_{i j}$ and $C_{i j k l}$ denote stress, strain, and fourth-order elastic tensor, respectively. $C_{i j k l}(\mathbf{x})$ is expressed in terms of local Young's modulus $E(\mathbf{x})$ and Poisson ratio $\nu(\mathbf{x})$ as
\begin{align*}
C_{ijkl}(\mathbf{x}) =\;& 
\frac{E(\mathbf{x}) \nu(\mathbf{x})}{(1 + \nu(\mathbf{x}))(1 - 2\nu(\mathbf{x}))} 
\, \delta_{ij} \delta_{kl} + \frac{E(\mathbf{x})}{2(1 + \nu(\mathbf{x}))} 
\left( \delta_{ik} \delta_{jl} + \delta_{il} \delta_{jk} \right),
\end{align*}
where $\delta_{i j}$ denotes the Kronecker delta function. $G_{i j k l}^{(0)}$ denotes Green's operator corresponding to the homogeneous reference material with elastic tensor $C_{i j k l}^{0}$ \cite{Moulinec1998FFT}. $\bar{\varepsilon}_{i j}$ denotes far field strain used to apply the strain-controlled boundary condition, equal to the average strain over the RVE domain $\Omega$.
In this example, we model the point-wise strain concentration tensor $A_{i j k l}(\mathbf{x}) \in \mathbb{R}^{3  \times 3} $, which linearly maps $\bar{\varepsilon}_{i j}$ to $\varepsilon_{i j}(\mathbf{x})$ independent of loading conditions:
\begin{align*}
    \varepsilon_{i j}(\mathbf{x})=A_{i j k l}(\mathbf{x}): \bar{\varepsilon}_{kl}.
\end{align*}
$A_{i j k l}(\mathbf{x})$ can be computed by triply solving ~\eqref{setup:dotLS} subject to three orthogonal unit far field strain $\overline{\varepsilon}_{ij}$ as
\begin{equation}
    \overline{\varepsilon}_{ij} = \begin{Bmatrix}[1,0,0]^{T},
[0,1,0]^{T},
[0,0,1]^{T}\end{Bmatrix}.
    \label{setup:macroBC}
\end{equation}

Due to the intrinsic symmetry of the RVE configuration, the resulting strain concentration tensor field $A_{i j k l}(\mathbf{x})$ exhibits a symmetric pattern with respect to both $x=\frac{1}{2}$ and $y=\frac{1}{2}$. {\color{black}
We aim to train a FunDiff model to reconstruct the strain concentration tensor field from sparse observations while respecting the inherent symmetry priors.
}
To this end. we modify the forward pass of the decoder to enforce these constraints. Let $\mathbf{z}$ denote the FAE encoder output latents. The symmetry-preserved decoder $\hat{\mathcal{D}}_{\theta}$ is then constructed as
\begin{align*}
\hat{\mathcal{D}}_\theta(\mathbf{z}, x, y) =\; 
& \mathcal{D}_\theta(\mathbf{z}, x, y) 
+ \mathcal{D}_\theta(\mathbf{z}, 1 - x, y)  + \mathcal{D}_\theta(\mathbf{z}, x, 1 - y) 
+ \mathcal{D}_\theta(\mathbf{z}, 1 - x, 1 - y).
\end{align*}
It can be easily shown that this formulation ensures that $\hat{\mathcal{D}}_\theta$ generates functions with perfect symmetry about the lines $x=1/2$ and $y=1/2$. Although here we only focus on such a relative simple symmetry for illustration purposes, it is worth noting here that more general symmetries can be embedded in a similar manner \cite{gao2024energy}.

\begin{figure*}
    \centering
    \includegraphics[width=0.95\linewidth]{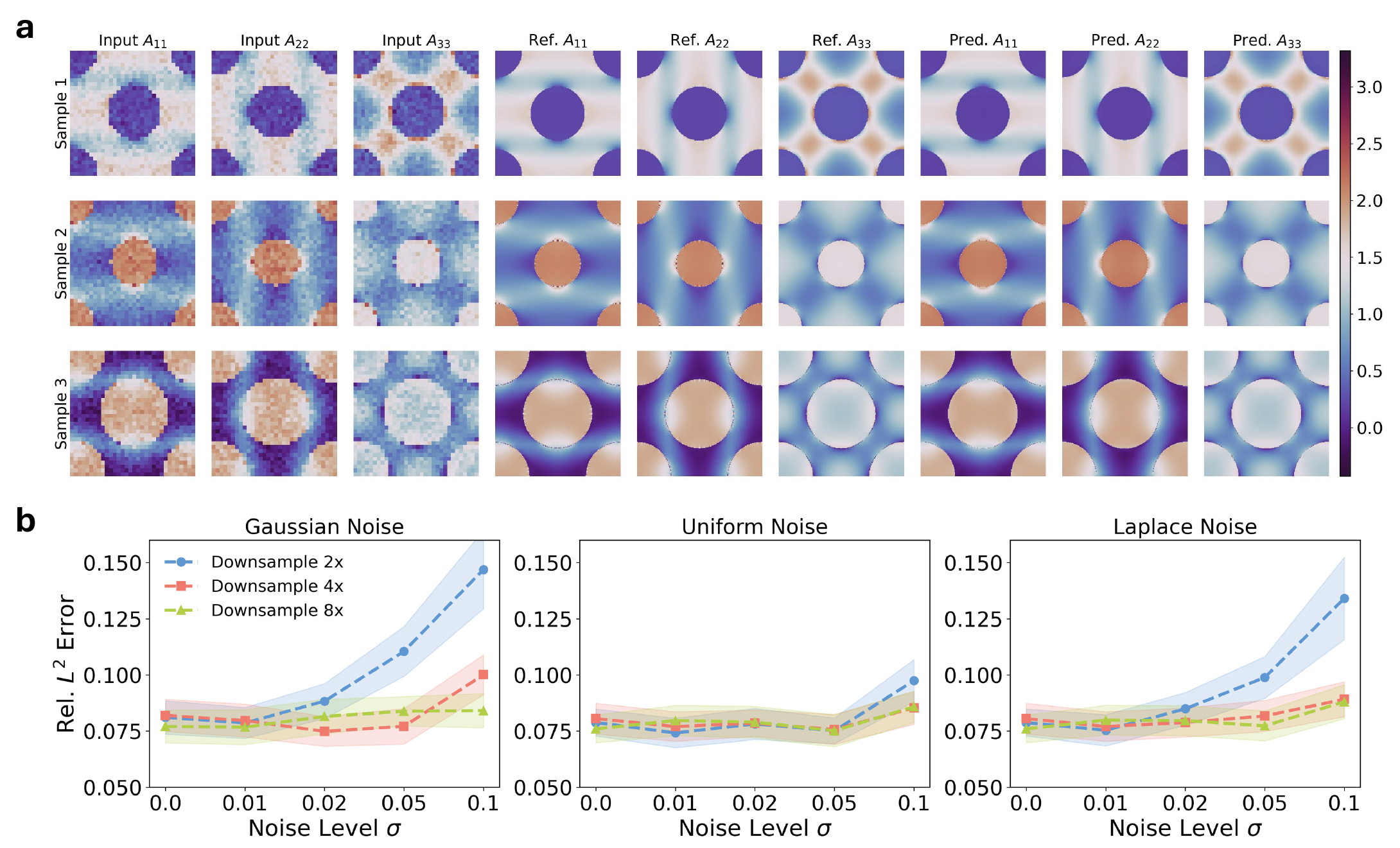}
    \caption{\textbf{Linear elasticity.} (\textbf{a}) Comparison of diagonal strain concentration tensor distributions (Voigt notation) between provided coarse measurements ($32 \times 32$) and reconstructions from the trained FunDiff model. The generated samples maintain enforced symmetry constraints and achieve a relative $L^2$ error of $5.12\%$ when compared to ground truth data at $256 \times 256$ resolution. (\textbf{b}) Comparison of reconstruction errors under different noise types and downsampling rates.}
    \label{fig:elasticity_combined}
\end{figure*}

Fig.~\ref{fig:elasticity_combined}a presents reconstructed samples generated by our FunDiff model from a coarse strain concentration field corrupted by 10\% Gaussian noise. The results show that our model preserves underlying symmetries and accurately resolves sharp material interfaces. To assess robustness against different types of noise, we report test errors in Fig.~\ref{fig:elasticity_combined}b. The consistently low errors across all noise configurations highlight the framework’s reliability in practical applications.

To further demonstrate the critical role of the continuous decoder in our framework, we implement a baseline using a discrete autoencoder with a Vision Transformer encoder and an MLP decoder projecting to the target pixel space. We train a diffusion transformer under identical conditions, with representative generated samples shown in Fig.~\ref{fig:elasticity_discrete}. Unlike our approach, the baseline fails to preserve symmetry or enforce physical continuity in strain fields, instead producing piecewise discontinuous predictions.

\subsection*{Turbulence mass transfer}

In the last example, we examine the model capacity of inferring non-intrusive quantities such as pressure in turbulent problems. Specifically, we consider turbulent mass transfer in a rectangular channel. This phenomenon plays a critical role in the chemical engineering process, directly influencing reaction kinetics, conversion efficiency, and product purity. Our study focuses on a computational domain $\Omega=\{(x, y) \mid 0.006<x<0.018,0<y<0.01\}$. {\color{black} Within this domain, we are interested in training a FunDiff model to infer the obstacle topology and pressure field given the sparse and noisy velocity measurements.}

The simulation employs a two-equation model for turbulent mass transfer based on the Reynolds-averaged Navier–Stokes (RANS) equations \cite{kou2025efficient}, which can be expressed as
\begin{align*}
    &\frac{\partial (\rho u_i)}{\partial x_i} = 0, \quad  \\
    &u_i \frac{\partial (\rho u_j)}{\partial x_i} = -\frac{\partial P}{\partial x_j} + \frac{\partial}{\partial x_i} \left[ \left( \mu_m + \mu_t \right) \frac{\partial u_j}{\partial x_i} \right], \quad  \\
    &u_i \frac{\partial (\rho C)}{\partial x_i} = \frac{\partial}{\partial x_i} \left[ \rho \left( D_m + D_t \right) \frac{\partial C}{\partial x_i} \right], \quad 
\end{align*}
where $p, u_i, x_i$ are the pressure, velocity and location respectively; the subscript $i$ represents the coordinate direction which can take $x$ and $y$ for the two-dimensional system. Here, $C$ is the time-averaged concentration; $\rho, \mu_m$ and $D_m$ are density, laminar viscosity, and laminar mass diffusivity, which are physical property constants and fixed to $1000 \mathrm{~kg} / \mathrm{m}^3, 1.003 \times 10^{-6} \mathrm{~m}^2 / \mathrm{s}, 1.003 \times 10^{-8} \mathrm{~m}^2 / \mathrm{s}$, respectively. $\mu_t$ and $D_t$ are the turbulent viscosity and turbulent mass diffusivity, respectively. 

We begin by training an FAE to reconstruct the key physical variables: velocity field, pressure field, and the signed distance function (SDF) representing obstacle geometry. To enhance the capacity of the model to handle inputs of varying resolutions, we augment training by randomly downsampling inputs by factors of 1, 2, and 4 while preserving the target resolution. 
The latent diffusion model is then trained with latent representations of the pressure and SDF,  conditioned on the encoded representations of the velocity field, obtained from the trained autoencoder's latent space. 

% \begin{figure}[H]
%     \centering
%     \includegraphics[width=1.0\linewidth]{figures/tmt/tmt_samples.png}
% \caption{{\em Turbulent Mass Transfer.} Left: Input velocity fields at low resolution with 10\% noise corruption. Middle: Comparison between reference and generated pressure fields. Right: Comparison between reference and generated obstacle geometry.}
%     \label{fig:tmt_samples}
% \end{figure}

% \begin{figure}[H]
%     \centering
%     \includegraphics[width=0.9\linewidth]{figures/tmt/tmt_error_comparison.pdf}
% \caption{{\em Turbulent Mass Transfer.} Relative $L^2$ error comparison between generated samples and numerical simulations across varying input resolutions and noise levels.}
%     \label{fig:tmt_error}
% \end{figure}

\begin{figure*}
    \centering
    \includegraphics[width=0.95\linewidth]{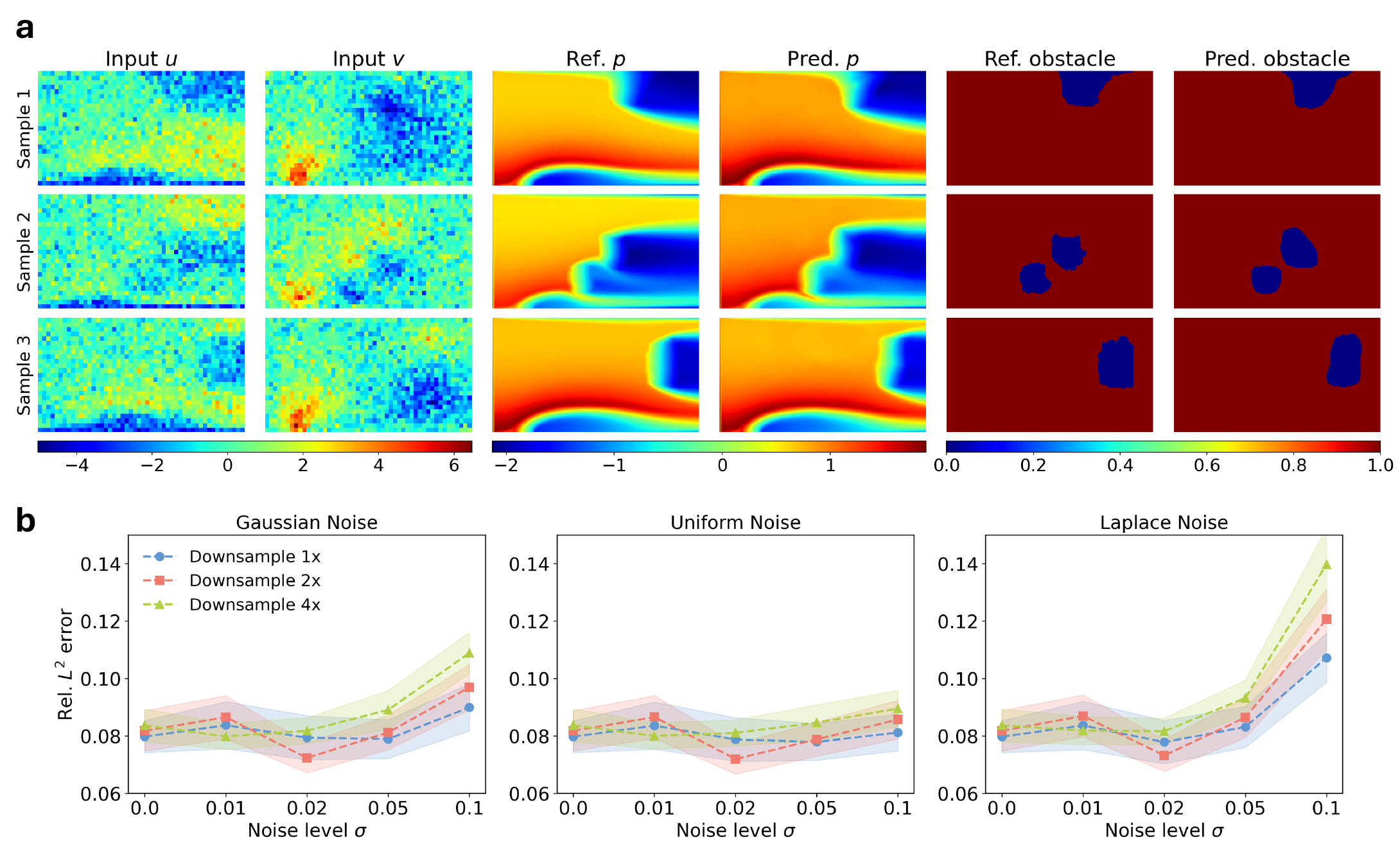}
    \caption{\textbf{Turbulent mass transfer.} (\textbf{a}) Left: Input velocity fields at low resolution with 10\% noise corruption. Middle: Comparison between reference and generated pressure fields. Right: Comparison between reference and generated obstacle geometry. (\textbf{b}) Relative $L^2$ error comparison between generated samples and reference solutions across varying input resolutions and noise levels.}
    \label{fig:tmt_combined}
\end{figure*}

Representative generated samples (Fig.~\ref{fig:tmt_combined}a) validate the model’s ability to infer pressure fields and reconstruct obstacle geometries despite heavily corrupted input velocity measurements. The reconstructions align closely with numerical estimations. Fig.~\ref{fig:tmt_combined}b presents the test error across various downsampling factors, noise levels, and noise types. The results indicate that our model remains robust even under severe noise corruption and low-resolution inputs. Our results demonstrate the model’s effectiveness in preserving critical flow features and geometric details, and highlight the framework’s potential for real-world applications where high-fidelity reconstructions are required from sparse or noisy measurements.

\section*{Discussion}
In this work, we propose FunDiff, a flexible and efficient framework for generative modeling in function spaces, addressing fundamental challenges in generating continuous function data while preserving physical constraints.
Our primary contributions are twofold: (1) a rigorous theoretical foundation establishing conditions for function distribution approximation via autoencoders, and (2) a novel constrained function autoencoder (FAE) that inherently enforces physical principles. Through extensive experimentation across diverse physical systems, we have demonstrated that our approach successfully generates physically valid samples while maintaining high fidelity to the target PDE solution distributions.

While our current work lays a strong foundation for physics-informed generative modeling, several promising directions for future research emerge. 
{\color{black}
One particularly interesting direction is to develop not purely data-driven, but physics-informed diffusion-based PDE solvers. 
This could be achieved by incorporating PDE residuals directly into the score function of diffusion models and by using additional initial and boundary conditions as guidance during sampling, thereby steering the generated samples toward the corresponding PDE solutions.
}
Another natural extension would be to adapt our framework to handle more complex geometries, thereby enabling broader applicability in real-world scientific and engineering problems. Handling complex domains necessitates incorporating geometric inductive biases, such as mesh-based representations, implicit neural surfaces, or graph-based function approximations. 
Moreover, extending our approach to multiphysics applications represents a critical and exciting frontier. Many engineering and scientific challenges involve the interplay of multiple physical processes, such as fluid-structure interactions, thermo-mechanical coupling, and electromagnetic-thermal phenomena, which require models that can simultaneously capture and reconcile diverse physical constraints. 
These advancements will significantly expand the scope of generative modeling for physical systems and ultimately contribute to more accurate and efficient simulation-based design across disciplines.

\section*{Methods}

To realize our theoretical framework for generative modeling in function spaces, we develop FunDiff, a practical implementation that addresses the key challenges of handling infinite-dimensional function distributions while preserving physical constraints. Our approach must accommodate discretized functions at varying resolutions, enable continuous evaluation at arbitrary coordinates, and flexibly incorporate domain-specific physical priors without compromising the generative modeling process. The architecture is designed to decouple physical constraint enforcement from diffusion training, ensuring stable optimization while maintaining mathematical rigor in constraint satisfaction. We detail the FunDiff network architecture, training procedures, and constraint integration mechanisms that enable robust function generation across diverse scientific domains.

\subsection*{Network architecture}

% To realize the framework of function generative models outlined in Section \ref{sec:fae-overview}, we propose FunDiff (Fig.~\ref{fig:pipeline}), which consists of a novel FAE and a latent diffusion model to generate continuous functions. The FAE learns the mapping from discrete function representations to continuous ones, while also enabling the enforcement of prior physical constraints.
% The latent diffusion model is then trained directly on the latent manifold produced by the pretrained FAE encoder.
% Our generation process follows a simple two-step procedure: we first sample latent codes by integrating the reverse-time ODE from Gaussian noise to the target data distribution, and then transform these codes into continuous functions via the FAE decoder. 

The FAE processes discretized functions $\mathbf{f} \in \mathbb{R}^{n \times d}$, where $n$ denotes the spatial discretization resolution and $d$ the codomain dimension. The FunDiff architecture comprises three core components: (1) a resolution-invariant encoder $\mathcal{E}$ combining a Vision Transformer (ViT) \cite{dosovitskiy2020image} and a Perceiver module \cite{jaegle2021perceiver}, (2) a latent diffusion model operating on the compact latent space, and (3) a physics-aware decoder $\mathcal{D}$ that reconstructs continuous functions.

\paragraph{Encoder.} 
The encoder $\mathcal{E}$ employs a ViT backbone to extract features and a Perceiver module to handle functions of different discretizations.
Input functions $\mathbf{f} \in \mathbb{R}^{H \times W \times C}$ are first processed by ViT-style patchification, resulting in flattened patches $\mathbf{f}_p \in \mathbb{R}^{L \times D}$, where $L=\frac{H}{P} \times \frac{W}{P}$ represents  the length of sequence, $P$ is the size of the patch, and $D$ denotes the embedding dimension.
To accommodate varying resolutions, trainable positional embeddings (PE) are dynamically interpolated to match the input sequence length $L$, yielding position-aware patch embeddings $\mathbf{f}_{p e}=\mathbf{f}_p+$ Interp$(\mathrm{PE}, L)$.

A perceiver-style cross-attention module maps these variable-length inputs to fixed-dimensional latent representations. Specifically, we learn a predefined number of latent input queries $\mathbf{z}_q \in \R^{N \times D}$. These queries serve as learnable parameters in a cross-attention module, which processes the visual features of our input as follows 
\begin{align*}
    \mathbf{z}^{\prime} & = {\mathbf{z}}_q + \operatorname{MHA}\left(\operatorname{LN}\left({\mathbf{z}_q} 
    \right), \operatorname{LN}\left(\mathbf{f}_{pe}
    \right), \operatorname{LN}\left(\mathbf{f}_{pe}
    \right)\right) , \\
\mathbf{z}_{agg} & = \mathbf{z}^{\prime} + \operatorname{MLP}\left(\operatorname{LN}\left(\mathbf{z}^{\prime}\right)\right),
\end{align*}
where MHA, LN, and MLP represent multi-head self-attention, layer normalization, and a multi-layer perceptron, respectively. We further process the aggregated tokens $\mathbf{z}_{agg}$ using a sequence of $L$ pre-norm Transformer blocks \cite{vaswani2017attention, xiong2020layer},
\begin{align*}
\mathbf{z}_{0} &= \operatorname{LN}(\mathbf{z}_{agg}), \\
\mathbf{z}_{\ell}^{\prime} & =\operatorname{MSA}\left(\operatorname{LN}\left(\mathbf{z}_{\ell-1}\right)\right)+\mathbf{z}_{\ell-1}, & & \ell=1, \dots, L, \\
\mathbf{z}_{\ell}, & =\operatorname{MLP}\left(\operatorname{LN}\left(\mathbf{z}_{\ell}^{\prime}\right)\right)+\mathbf{z}_{\ell}^{\prime}, & & \ell=1, \dots, L,
\end{align*}
where MSA is a standard multi-head self-attention module used to model long-range dependencies across the sequence.

\paragraph{Decoder.}
The decoder module draws inspiration from the Continuous Vision Transformer (CViT) \cite{wang2024bridging}, which
enables continuous evaluation at any query coordinate through a cross-attention mechanism between query coordinates and encoded features. We start by embedding coordinates into the latent space, which can be efficiently achieved through either random Fourier features \cite{tancik2020fourier} or grid-based embeddings \cite{wang2024bridging}.

The core of our decoder uses embedded coordinates $\mathbf{x}_{0} \in \R^{D}$ as queries, with the encoder output $\mathbf{z}_L$ serving as both keys and values in a series of $K$ cross-attention Transformer blocks:
\begin{align*}
    \mathbf{x}^{\prime}_{k} & = \mathbf{x}_{k - 1} + \operatorname{MHA}\left(\operatorname{LN}\left(\mathbf{x}_{k - 1} 
    \right), \operatorname{LN}\left(\mathbf{z}_L
    \right), \operatorname{LN}\left(\mathbf{z}_L
    \right)\right), & & k=1 \ldots K, \\
\mathbf{x}_{k} & =\mathbf{x}^{\prime}_{k} + \operatorname{MLP}\left(\operatorname{LN}\left( \mathbf{x}^{\prime}_{k}\right)\right), & & k=1 \ldots K.
% \mathbf{y}_{K+1} &= \operatorname{LN}\left(\mathbf{y}_K
%     \right)
\end{align*}

Finally, a small feedforward network projects the features $\mathbf{x}_K$ to the desired output dimension. This architecture naturally extends to other function representations (e.g., point clouds via PointNet \cite{qi2017pointnet} or meshes via graph neural networks \cite{pfaff2020learning}) through appropriate input processing modules.

\paragraph{Latent diffusion with conditional Diffusion Transformer.}
We implement the diffusion process in the learned latent space using a Diffusion Transformer (DiT)~\cite{peebles2023scalable}. We condition the model on coarse solution measurements $\mathbf{c}$ by encoding them through the pretrained FAE encoder $\mathcal{E}_\theta$. The resulting latent-space guidance vectors are additively combined with the diffusion inputs:
\begin{equation*}
\tilde{ \mathbf{z}}_1 = \mathbf{z}_1 + \mathcal{E}_\theta(\mathbf{c}).
\end{equation*}

The DiT backbone consists of a sequence of Transformer blocks that utilize AdaLN-Zero modulation \cite{peebles2023scalable}, an enhanced variant of Adaptive Layer Normalization designed for stable time-dependent conditioning. For each block, the modulation parameters are computed as
\begin{align*}
\mathbf{\gamma}, \mathbf{\beta} &= \text{MLP}(t), \\
\mathbf{z}' &= \text{LN}(\mathbf{z}) \odot \mathbf{\gamma} + \mathbf{\beta},
\end{align*}
where $t$ represents the timestep. Following the approach in Ref.~\cite{peebles2023scalable}, we incorporate element-wise scaling factors $\mathbf{\alpha}$, initialized to zero, in the residual pathways:
\begin{equation*}
\mathbf{z}_{\text{out}} = \mathbf{\alpha} \odot \text{MSA}(\mathbf{z}') + \mathbf{z}.
\end{equation*}
By initializing these scaling factors to zero, each Transformer block initially acts as an identity transform. This strategy enables gradual learning of temporal and conditional modulations while maintaining stable gradient flow during the early stages of training.

\paragraph{Training.}

Our training procedure consists of two stages. In the first stage, we train an FAE to learn a compact and expressive latent representation of functions. We consider a set of ground truth functions $\left\{f_i\right\}_{i=1}^N$, where each function $f_i: \mathcal{X} \rightarrow \mathbb{R}$ maps input coordinates $\mathbf{x} \in \mathcal{X} \subset \mathbb{R}^d$ to scalar outputs. For training, each function $f_i$ is discretized at a fixed set of input locations $\left\{\mathbf{x}_j\right\}_{j=1}^M$, resulting in a vector of evaluations $\mathbf{f}_i=\left[f_i\left(\mathbf{x}_1\right), \ldots, f_i\left(\mathbf{x}_M\right)\right] \in \mathbb{R}^M$. The reconstruction loss is given by
\begin{align*}
\mathcal{L}_\text{FAE} = \frac{1}{N M} \sum_{i=1}^N \sum_{j=1}^M\left| f_i(\mathbf{x}_j) - \mathcal{D}(\mathcal{E}(\mathbf{f}_i))(\mathbf{x}_j) \right|^2,
\end{align*}
where $\mathcal{E}$ and $\mathcal{D}$ denote the encoder and decoder respectively, and $f_i(\mathbf{x}_j)$ represents the value of a ground truth function $f_i$ at query point $\mathbf{x}_j$. During training, we randomly sample functions $\{f_i\}$ and query coordinates $\{\mathbf{x}_j\}_{j=1}^M$ at each iteration.

In the second stage, we freeze the pre-trained FAE and train the DiT model in the induced latent space. Following the rectified flow framework, the training objective is formulated as
\begin{align}
\label{eq:diffusion_loss}
\mathcal{L}_{\text{DiT}} &= \int_0^1 \mathbb{E}_{\mathbf{f} \sim p_{\text{data}}}\left[\left|\left(\mathbf{z}_1-\mathbf{z}_0\right)-\mathbf{g}(\mathbf{z}_\tau, \tau)\right|^2\right] d\tau, \
&\text{with} \quad \mathbf{z}_\tau = (1 - \tau)\mathbf{z}_1 + \tau \mathbf{z}_0, \quad \mathbf{z}_1 = \mathcal{E}(\mathbf{f}),
\end{align}
where $\mathbf{z}_1$ is obtained by encoding the input function through the pre-trained FAE encoder, $\mathbf{z}_0 \sim \mathcal{N}(0, \mathbf{I})$ is sampled from the standard normal distribution, and $\mathbf{g}(\mathbf{z}_t, t)$ is parameterized by our DiT model predicting the velocity field. In practice, we approximate the time integral by uniformly sampling $\tau \sim \mathcal{U}[0,1]$ at each training step.

\paragraph{Inference.}

At inference, we follow a straightforward two-step generation process. First, we sample latent codes from our learned diffusion model by solving the ODE using a standard ODE solver (e.g., Runge-Kutta method):
\begin{align*}
\frac{d\mathbf{z}(t)}{dt} = \mathbf{g}(\mathbf{z}(t), t), \quad t \in [0,1],
\end{align*}
starting from a sample of standard Gaussian noise $\mathbf{z}_1 \sim \mathcal{N}(0, \mathbf{I})$ and integrating backward to obtain $\mathbf{z}_0$. This process is deterministic and typically requires 20--50 function evaluations for high-quality samples. Second, we decode the sampled latent codes through the FAE decoder to obtain the final continuous function. Importantly, since our decoder operates on arbitrary query points, we can evaluate the generated function at any desired resolution or location in the domain.

\paragraph{Implementation details.}
We maintain consistent architectural choices and training procedures across all experiments unless explicitly stated otherwise. Our FAE architecture comprises an 8-layer encoder and a 4-layer decoder, both utilizing an embedding dimension of 256 and MLP width of 512. The encoder's patch size is adaptively selected based on input resolution. The DiT employs 8 Transformer blocks with an embedding dimension of 256 and MLP width of 512. All attention operations use 8 heads.

\paragraph{Training protocol.}
For the FAE training, we employ the AdamW optimizer~\cite{kingma2014adam} with a weight decay of $10^{-5}$. The learning rate follows a two-phase schedule: an initial linear warm-up from zero to $10^{-3}$ over 2,000 steps, followed by exponential decay with a factor of 0.9 every 2,000 steps. Each training batch contains 16 samples, with 4,096 query coordinates and their corresponding outputs randomly sampled from the grid. To enhance the model's resolution invariance,  we randomly downsample inputs by different factors (e.g., 1, 2, 4, 8) during training while optimizing the model to reconstruct the original high-resolution input function by minimizing the MSE loss.

The DiT follows a similar optimization strategy but operates with larger batches of 128 samples over $10^5$ iterations. For training the conditioned diffusion models, we generate conditions by randomly downsampling input functions and encoding them through the pretrained FAE encoder.
Detailed hyperparameters for all experiments are provided in Appendix \ref{appendix: experiments}.

\bibliography{sample}

% \noindent LaTeX formats citations and references automatically using the bibliography records in your .bib file, which you can edit via the project menu. Use the cite command for an inline citation, e.g.  \cite{Hao:gidmaps:2014}.

% For data citations of datasets uploaded to e.g. \emph{figshare}, please use the \verb|howpublished| option in the bib entry to specify the platform and the link, as in the \verb|Hao:gidmaps:2014| example in the sample bibliography file.

\section*{Acknowledgments}

LL acknowledges support from the U.S. Department of Energy, Office of Advanced Scientific Computing Research, under Grants No. DE-SC0025593 and No. DE-SC0025592, and the U.S. National Science Foundation under Grants No. DMS-2347833 and No.~DMS-2527294. SW acknowledges support from the NVIDIA Academic Grant Program. We thank Dr.\ Chenhui Kou for providing the turbulence mass transfer dataset and Dr.\ Mingyuan Zhou for insightful discussions.

\section*{Author contributions statement}
S.W., Z.D. and L.L. conceived the project and developed the main methodology. S.W. implemented the core algorithms and conducted the experiments. Z.D. contributed to the theoretical analysis. T.L. and S.S. assisted with numerical simulation. L.L. supervised the project and provided critical revisions. All authors discussed the results and contributed to writing the manuscript.

% \section*{Additional information}

% To include, in this order: \textbf{Accession codes} (where applicable); \textbf{Competing interests} (mandatory statement). 

% The corresponding author is responsible for submitting a \href{http://www.nature.com/srep/policies/index.html#competing}{competing interests statement} on behalf of all authors of the paper. This statement must be included in the submitted article file.

\appendix
\input{appendix}

\end{document}

%% file: appendix.tex
\renewcommand{\thesection}{S\arabic{section}}
\renewcommand{\thesubsection}{\thesection.\arabic{subsection}}
\renewcommand{\thefigure}{S\arabic{figure}}
\renewcommand{\thetable}{S\arabic{table}}
\renewcommand{\theequation}{S\arabic{equation}}
\renewcommand{\thetheorem}{S\arabic{theorem}}
\setcounter{figure}{0}
\setcounter{table}{0}
\setcounter{equation}{0}
\setcounter{theorem}{0}

\clearpage
\section{Notations}
\label{app:notations}

Our notations are summarized in Table \ref{tab:notation}.

\begin{table}[htbp]
\centering
\renewcommand{\arraystretch}{1.2}
\caption{\textbf{Main symbols and notations used in this work.}}
\begin{tabular}{|>{\centering\arraybackslash}m{0.2\textwidth}|>{\arraybackslash}m{0.7\textwidth}|}
\hline
\rowcolor{gray!30}
\textbf{Notation} & \textbf{Description} \\ \hline
% \multicolumn{2}{|>{\columncolor{gray!15}}c|}{\textbf{Operator Learning}} \\ \hline
% $\mathcal{X}$ & The input function space \\ 
% $\mathcal{Y}$ & The output function space \\ 
% $f \in \mathcal{X}$ & Input function 
% \\
% $s \in \mathcal{Y}$ & Output function  \\
% $y$ & Query coordinate in the input domain of $s$ \\
% $\mathcal{G}: \mathcal{X} \rightarrow \mathcal{Y}$ & The operator mapping between function spaces 
% \\
% $\curlyE: \curlyX \to \R^n$ & Encoder mapping \\
% $\curlyD: \R^n \to \curlyY $ & Decoder mapping \\
% \hline
\multicolumn{2}{|>{\columncolor{gray!15}}c|}{\textbf{FunDiff}} \\
\hline  
 \text{MLP}  & $\text{Multilayer perceptron}$    \\ 
$\text{PE}$ &  Positional embedding \\ 
% $\text{SA}$ & Self-attention \\
$\text{MSA}$ & Multi-head self-attention \\
$\text{MHA}$ & Multi-head attention \\
% $\text{CSA}$ & Multi-head cross-attention \\
$\text{LN}$ & Layer normalization \\ 
$P$ & Patch size of Vision Transformer \\
$D$ & Embedding dimension of Vision Transformer \\
% $N_x \times N_y$ & Resolution of dummy grid  \\
\hline
\multicolumn{2}{|>{\columncolor{gray!15}}c|}{\textbf{Partial differential equations}} \\ \hline
$\boldsymbol{\sigma}$ & Stress vector \\
$\mathbb{C}$ & Fourth-order elastic tensor \\
$\mathbb{A}$ & Strain concentration tensor \\
$E(x)$, $\nu(x)$ & A pair of Young's modulus and Poisson ratio\\
$\mathbf{u}$ & Velocity field  \\
$w$  & Voriticty field \\ 
$p$ & Pressure field  \\
% $\mathbf{f}$ & Buoyancy  \\
$\nu$ & Viscosity \\ 
\hline
\multicolumn{2}{|>{\columncolor{gray!15}}c|}{\textbf{Hyperparameters}} \\ \hline
$B$ & Batch size \\ 
$Q$ & Number of query coordinates in each batch \\ 
$D$ & Number of latent variables of interest \\
$H \times W$ & Resolution of spatial discretization \\
 \hline
\end{tabular}
\label{tab:notation}
\end{table}

We present detailed mathematical definitions as follows.
The Wasserstein-1 distance (also known as the Earth Mover's Distance) between two probability measures $\mu$ and $\nu$ on a metric space $(\mathcal{X}, d)$ is defined as
\[
W_1(\mu, \nu) = \inf_{\gamma \in \Pi(\mu, \nu)} \mathbb{E}_{(X,Y) \sim \gamma} [d(X,Y)],
\]
where:
\begin{itemize}
\item $d(\cdot, \cdot)$ is the given metric (e.g., Euclidean distance) that measures the cost of transporting mass from $X$ to $Y$.
\item $\Pi(\mu, \nu)$ is the set of \textit{couplings} of $\mu$ and $\nu$, i.e., joint probability distributions $\gamma$ on $\mathcal{X} \times \mathcal{X}$ whose marginals are $\mu$ and $\nu$:
\[
\gamma(A \times \mathcal{X}) = \mu(A), \quad \gamma(\mathcal{X} \times B) = \nu(B) \quad \text{for all measurable sets } A, B \subset \mathcal{X}.
\]
\end{itemize}
Thus, $W_1(\mu, \nu)$ represents the minimal expected cost to transport the mass of $\mu$ to match $\nu$, given that the cost function is the distance $d(x, y)$. In a dual form, $W_1$ distance can be equivalently expressed as
\[W_1(\mu, \nu) = \sup_{f\text{~is 1-Lipschitz}} \left[\int_{\mathcal X} f \,\rd\mu - \int_{\mathcal X} f \,\rd\nu \right]. \]

Another notation we introduce here is the pushforward measure.  Given two measurable spaces $(\mathcal{X}, \mathcal{A})$ and $(\mathcal{Y}, \mathcal{B})$, as well as a measurable function $T: \mathcal{X} \to \mathcal{Y}$, the pushforward measure $T_{\#} \mu$ of a measure $\mu$ on $(\mathcal{X}, \mathcal{A})$ is a measure on $(\mathcal{Y}, \mathcal{B})$ defined by
\[
(T_{\sharp} \mu)(B) = \mu(T^{-1}(B)), \quad \forall B \in \mathcal{B}.
\]
This means that for any measurable set $B \subset \mathcal{Y}$, the pushforward measure assigns a value based on how much mass $\mu$ assigns to its preimage $T^{-1}(B)$ in $\mathcal{X}$. In other words, $T_{\sharp}\mu$ is the underlying distribution of $T(x)$ where $x\sim \mu$. 

Besides, for two functions $f(x), g(x) > 0$, we write $f\lesssim g$ if there exists a positive constant $C > 0$ such that $f(x)\leqslant C\cdot g(x)$ holds for sufficiently large $x > 0$. If $f \lesssim g$ and $g\lesssim f$ hold at the same time, then we denote it as $f\asymp g$, which means that there exist positive constants $c, C > 0$ such that $c\cdot g(x) \leqslant f(x)\leqslant C\cdot g(x)$ for sufficiently large $x > 0$.

\clearpage
\section{Preliminaries}
\label{app: Preliminaries}
Before introducing our generative models in function spaces, we provide some basic knowledge of diffusion models and its Minimax optimality. 

We begin by reviewing foundational concepts of diffusion models in both discrete and continuous settings. Diffusion models form a class of generative frameworks that approximate complex data distributions through a two-phase process: a forward noising process that incrementally corrupts the data, followed by a backward denoising process that reconstructs data from noise.

In the discrete setting \cite{sohl2015deep,ho2020denoising}, the forward process is modeled as a first-order Markov chain with $N$ steps:
\begin{align}
q(\mathbf{x}_{1:N} \mid \mathbf{x}_0) 
&= \prod_{k=1}^N q(\mathbf{x}_k \mid \mathbf{x}_{k-1}), \\
q(\mathbf{x}_k \mid \mathbf{x}_{k-1}) 
&= \mathcal{N}\bigl(a_k \mathbf{x}_{k-1}, \, b_k^2 \mathbf{I}\bigr).
\end{align}
where $\bx_0 \in \mathbb{R}^d$ is the original data point, $\bx_k \in \mathbb{R}^d$ denotes the noised data at discrete time step $k$, and the coefficients $a_k \in \mathbb{R}$ and $b_k > 0$ are predefined scalars that control signal decay and noise level at step $k$. The forward process incrementally injects Gaussian noise while decaying the signal, and the Gaussian structure ensures that the marginal distribution $q(\bx_k \mid \bx_0)$ remains analytically tractable for all $k$. This facilitates both efficient sampling and likelihood estimation. During the backward process, a neural network $\vs_{\bm{\theta}}(\bx_k, t_k)$ is trained to approximate the score function $\nabla_{\bx_k} \log q(\bx_k)$, where $\bm{\theta}$ are the learnable network parameters and $t_k$ is a scalar or embedding indicating the diffusion timestep associated with $\bx_k$. The estimated score function is then used to reverse the noising process, enabling the generation of new samples from noise via stochastic denoising steps.

While classical diffusion models rely on stochastic differential equations (SDEs) or discrete Markov chains to reconstruct samples iteratively, rectified flow provides an appealing deterministic alternative. Instead of adding and removing noise through a stochastic process, rectified flow formulates the generative task as a deterministic transport problem, where a time-dependent vector field $\vv(\bx, t)$ is learned to smoothly morph a simple base distribution (e.g., standard Gaussian) into the data distribution \cite{liu2209rectified}. Specifically, rectified flow defines a continuous evolution via the ordinary differential equation (ODE):
\begin{align*}
\frac{\mathrm{d} \bx}{\mathrm{d} t} = \vv(\bx, t),
\end{align*}
where $\bx \in \mathbb{R}^d$ is the evolving sample at time $t \in [0,1]$ and $\vv: \mathbb{R}^d \times [0,1] \to \mathbb{R}^d$ is a learnable vector field optimized to carry mass from the source to the target distribution. In many cases, $\vv(\bx, t)$ is parameterized as the gradient of a scalar potential function, i.e., $\vv(\bx, t) = \nabla_{\bx} \phi(\bx, t)$, encouraging samples to travel along energy-efficient paths. This approach is rooted in the theory of optimal transport, aiming to minimize distortion and ensure that the learned trajectories follow nearly straight-line flows in high-dimensional space. The result is a computationally efficient and conceptually elegant framework for generative modeling.

To combine the advantages of deterministic flows and score-based models, recent work employs flow matching objectives. These objectives aim to learn a transport vector field by matching it to a known or estimated flow—such as the one induced by the diffusion model’s score function—via a regression loss. In this sense, flow matching acts as a unifying framework that bridges stochastic diffusion models and deterministic flow-based models. Importantly, rectified flow can be viewed as a special case of flow matching, where the target flow is derived from a score-based process and then refined into a smooth ODE trajectory. Notably, the $c$-rectified flow \cite{liu2209rectified} extends this idea by introducing corrective updates to the vector field, refining the transport map through multiple iterations and improving convergence to the optimal transformation \cite{villani2009optimal, caffarelli1996boundary}.

From this viewpoint, score-based diffusion models can be seen as stochastic analogs of deterministic flow models. Indeed, the backward process in a discrete diffusion model often takes the form
\begin{align*}
p_{\bm{\theta}}(\bx_{k-1} \mid \bx_k) = \mathcal{N}\left(u_k \hat{\bx}_0(\bx_k) + v_k \vs_{\bm{\theta}}(\bx_k, t_k), w_k^2 \bm{I}\right),
\end{align*}
where $\hat{\bx}_0(\bx_k)$ is an estimate of the original data sample conditioned on $\bx_k$, $u_k, v_k \in \mathbb{R}$ are scalar coefficients derived from the noise schedule, $w_k^2 > 0$ is the reverse variance at step $k$ and $\vs_{\bm{\theta}}(\bx_k, t_k)$ stands for the trained score function. As the variance $w_k^2$ approaches zero, this stochastic update converges to a deterministic transformation, recovering the structure of a flow model. This connection reveals that diffusion models, flow matching, and rectified flow all inhabit a shared conceptual space—each providing different approximations to the same underlying transport dynamics between probability distributions.

Unlike classical variance-preserving diffusion models \cite{ho2020denoising}, which depend on carefully designed noise schedules for both forward and reverse processes, rectified flow directly parameterizes the dynamics through a deterministic ODE. This not only simplifies the learning pipeline but also improves inference efficiency by eliminating the need for repeated noise sampling. Despite its deterministic nature, rectified flow retains strong expressiveness and sample quality, and its connection to score matching and optimal transport principles makes it a compelling direction for scalable generative modeling.

\subsection*{Density estimation and the minimaxity of diffusion models}

As a core problem in statistics and machine learning, the purpose of density estimation is to solve the following problem: given $n$ i.i.d. data points $\{x_i\}_{i=1}^n$ sampled from an unknown distribution $p^*$, we want to construct an approximate distribution $\hat{p}$ based on the $n$ given data points. Density estimation arises in diverse applications, from anomaly detection and clustering to modern generative modeling, where having an accurate handle on the underlying distribution is crucial. One broad class of solutions is parametric methods, which assume that the ground truth distribution $p^*$ lies within a known family of distributions $\{p_{\theta}:~\theta\in \Theta\}$. The density estimation problem then reduces to finding the optimal parametric vector $\hat{\theta}$ via log-likelihood maximization:
\[\hat{\theta} = \arg\max_{\theta\in\Theta} \sum_{i=1}^n \log p_{\theta}(x_i).\]
If $p^*$ aligns well with the chosen family, parametric models can be computationally efficient and produce robust estimates. However, when data exhibit complex or multimodal patterns that deviate from the assumed structure, parametric methods can underfit or produce biased results. In contrast, non-parametric approaches place fewer constraints on the distribution’s form. Kernel density estimation (KDE) is one of the most common examples:
\[\hat{p}_h(x) = \frac{1}{nh^d} \sum_{i=1}^n K\left(\frac{x-x_i}{h}\right).\]
Here, $K(\cdot)$ is a kernel function (which is usually chosen as Gaussian), $h$ is the bandwidth, and $d$ is the input dimension. Because nonparametric methods can model complex structures directly from data, they often uncover finer distributional details than simpler parametric models. However, as dimensionality increases, these methods face significant challenges, commonly referred to as the curse of dimensionality, leading to high sample complexity and potentially prohibitive computational costs.

Recently, diffusion models have emerged as powerful generative approaches that approximate the data distribution by reversing a forward noise-adding process. Although widely recognized for their empirical success, these models also admit a theoretical interpretation under the lens of minimax optimality for density estimation. In fact, under suitable assumptions on the function space that characterizes the ground truth distribution $p^*$, one can show that diffusion-based estimators achieve minimax rates for distribution (or score) recovery. Here we outline the theoretical underpinnings of this minimax perspective by first presenting a lower bound result (Theorem \ref{diffusion-lower}) that establishes a fundamental limit for density estimation in the space of all continuous functions $\mathcal C(\bR^d)$, and then provide an upper bound result (Theorem \ref{diffusion-upper}) that shows how diffusion-based methods (and related estimators) can achieve this rate, thus elucidating the minimax guarantees behind modern generative modeling frameworks.

\begin{theorem}[Lower Bound, Theorem 3 of Ref.~\cite{niles2022minimax}]
\label{diffusion-lower}
Let $\alpha \geqslant 0$ denote the smoothness parameter, $d\geqslant 1$ the dimension, and $c>0$ a positive constant. Suppose we consider the class of densities $\mathcal F \subseteq \mathcal C^{\alpha}(\bR^d)$ satisfying
\begin{itemize}
    \item[(i)] Smoothness: each density function $f\in \mathcal F$ has smoothness of order $\alpha$. 
    \item[(ii)] Lower bound: $f(x) \geqslant c$ for all $x\in \mathrm{supp}(f)$, which means the density is lower bounded by a positive constant in its support set.  
\end{itemize}
Then the estimation rate in the Wasserstein-1 distance satisfies the following lower bound given $n$ samples $\mu_1, \mu_2, \ldots, \mu_n\sim f$:
\begin{equation*}
\inf_{\hat{f}} \sup_{f\in \mathcal F} \mathbb E_f \left[W_1\left(\hat{f}, f\right)\right] \gtrsim \begin{cases} n^{-\frac{\alpha+1}{2\alpha+d}}, & d\geqslant 2, \\
n^{-1/2}, & d = 1,
\end{cases}
\end{equation*}
where the infimum is taken over all estimators $\hat{f}$ based on the sample of size $n$. 
\end{theorem}

\begin{theorem}[Upper Bound, Section 3.3 of Ref.~\cite{dou2024optimal}]
\label{diffusion-upper}
Let $\alpha \geqslant 0$ denote the smoothness parameter, $d\geqslant 1$ the dimension, and $c>0$ a positive constant. We still consider the class of densities $\mathcal F \subseteq \mathcal C^{\alpha}(\bR^d)$ defined in Theorem \ref{diffusion-lower}. By solving the backward diffusion SDE
\[\rd Y_t = \hat{s}(Y_t, T-t) \rd t + \rd W_t\]
with a non-parametric score estimator $\hat{s}(\cdot, t)$ injected, we achieve a density estimator $\hat{f}$ such that
\[\sup_{f\in\mathcal F} \mathbb E_f \left[W_1\left(\hat{f}, f\right)\right] \lesssim \begin{cases} n^{-\frac{\alpha+1}{2\alpha+d}}, & d > 2, \\ n^{-1/2}\log n, & d = 2, \\
n^{-1/2}, & d = 1.
\end{cases} \]
From these two theorems, we can see that score-based diffusion models are able to achieve minimax rate over the Wasserstein distance with respect to density estimation. 
\end{theorem}

\clearpage
\section{Theoretical proofs}
In this section, we provide theoretical proofs of our theorems, lemmas, and propositions.

\subsection{Density estimation error decomposition}
\label{sec:B.1}
In this part, we prove the decomposition of density estimation Wasserstein error in ~\eqref{eqn: framework}. According to the FAE framework, we have $\widehat{P} = G_{\sharp} \hat{p}$, which leads to 
\begin{equation*}
\begin{aligned}
W_1(P, \widehat{P}) &= W_1(P, G_{\sharp} \hat{p}) \overset{(a)}{\leqslant} W_1(P, G_{\sharp} p) + W_1(G_{\sharp} p, G_{\sharp} \hat{p})\\
&\overset{(b)}{\leqslant} \bE_{f\sim P} \|f - G\circ E(f)\|_2 + \mathrm{Lip}(G) \cdot W_1(p, \widehat{p}).
\end{aligned}
\end{equation*}
Here, $(a)$ is the direct result of the triangle inequality. $(b)$ holds because of the following two inequalities:
\[W_1(P, G_{\sharp} p) = W_1(P, G_{\sharp} E_{\sharp} P) = \sup_{\mathrm{Lip}(\mathcal L) \leqslant 1} \left[\bE_{f\sim P} \mathcal L(f) - \bE_{f\sim P} \mathcal L(G\circ E (f))\right] \leqslant \bE_{f\sim P} \|f-G\circ E(f)\|_2, \]
\[W_1(G_{\sharp} p, G_{\sharp} \hat{p}) \leqslant \sup_{\theta, \theta'} \frac{\|G(\theta)-G(\theta')\|_2}{\|\theta-\theta'\|_2} \cdot W_1(p, \hat{p}) = \mathrm{Lip}(G)\cdot W_1(p, \hat{p}). \]
Then it comes to our conclusion. 

\subsection{Proof of Theorem \ref{thm: 1}}
\label{sec:proof.1}
For any function $f\in \mathcal B$, denote its eigen-decomposition as $f = \sum_{i=1}^{\infty} \theta_i\cdot \sqrt{\mu_i}\psi_i$, then the reconstructed function
\[G\circ E(f) = \sum_{i=1}^{D} \theta_i\cdot \sqrt{\mu_i}\psi_i. \]
The reconstruction loss for $f\in\mathcal B$
\begin{equation*}
\Big\|f-G\circ E(f)\Big\|_2^2 = \sum_{i=1}^D \theta_i^2\cdot \mu_i \leqslant \mu_D \cdot \sum_{i=1}^D \theta_i^2 \leqslant \mu_D \cdot \sum_{i=1}^{\infty} \theta_i^2 = \mu_D \cdot \|f\|_\mH^2 \leqslant \mu_D, 
\end{equation*}
which leads to 
\[\bE_{f\sim P} \|f-G\circ E(f)\|_2 \leqslant \sqrt{\mu_D}.\]
For the next term $\mathrm{Lip}(G)$, we consider two vectors $\theta, \theta' \in \R^D$. Denote $f, f'\in \mathcal B$ as $f=G(\theta)=\sum_{i=1}^{D}\theta_i\cdot \sqrt{\mu_i}\psi_i$ and $f'=G(\theta')=\sum_{i=1}^{D}\theta_i'\cdot \sqrt{\mu_i}\psi_i$, then
\[\|f-f'\|_2^2 = \sum_{i=1}^D (\theta_i-\theta_i')^2 \cdot \mu_i \leqslant \mu_1 \cdot \|\theta-\theta'\|^2, \]
and we obtain that $\mathrm{Lip}(G) = \sqrt{\mu_1}$. Given the oracle loss $W_1(p, \hat{p}) \lesssim \varepsilon(D, n) = n^{-1/D}$, we have 
\[W_1(P,\widehat{P})\leqslant \sqrt{\mu_D} + \sqrt{\mu_1} \cdot n^{-1/D}. \]
Under the polynomial decay setting:
\[W_1(P, \widehat{P}) \lesssim n^{-1/D}+D^{-\beta}.\]
When the latent dimension is chosen as $D =\frac{1}{\beta}\cdot \frac{\log n}{\log\log n}$, we obtain that
\[W_1(P, \widehat{P}) \lesssim \left(\frac{\log\log n}{\log n}\right)^{\beta}.\]
Under the exponential decay setting:
\[W_1(P, \widehat{P}) \lesssim n^{-1/D} + \exp(-C_1/2 \cdot D^{\gamma}).\]
When we choose the optimal dimension $D = \left(\frac{2\log n}{C_1}\right)^{\frac{1}{1+\gamma}}$, we obtain that
\[W_1(P, \widehat{P}) \lesssim \exp\left(-(C_1/2)^{\frac{1}{1+\gamma}}(\log n)^{\frac{\gamma}{1+\gamma}}\right). \]

\subsection{Proof of Theorem \ref{thm: 2}}
\label{sec:proof.2}
In this proof, we still use the density loss decomposition. For the representation loss, we already have
\[\|f - G\circ E(f)\|_2 = \|f - f(\tilde{\theta}; \bx)\|_2 \leqslant \sqrt{\bE_{\bx\sim\mathrm{Unif}(\Omega)}|f(\bx)-f(\tilde{\theta}; \bx)|^{2}}\leqslant \frac{4}{\sqrt{m}}\]
holds for $\forall f\in\mathcal F_s$ according to the approximation error in Lemma \ref{lemma: approx-Barron}, which leads to
\[\bE_{f\sim P} \|f - G\circ E(f)\|_2 \leqslant \frac{4}{\sqrt{m}}. \]
For the next term $\mathrm{Lip}(G)$, we can prove the following result.
\begin{lemma}
\label{lemma: lipschitz}
For any two parameter sets $(a_k, \tb_k, c_k)_{k\in [m]}$ and $(a_k', \tb_k', c_k')_{k\in [m]}$ where $|a_k|, |a_k'|\leqslant\frac2m$, $\|\tb_k\|_1= \|\tb_k'\|_1 = 1$ and $|c_k|, |c_k'|\leqslant 1$ for $\forall k\in [m]$, then for $\theta = (\frac{m}{2}a_k, \tb_k, c_k)$ and $\theta' = (\frac{m}{2}a_k', \tb_k', c_k')$, it holds that
\[\|f(\theta; \cdot) - f(\theta'; \cdot)\|_2 \lesssim  \sqrt{\frac{d}{m}} \cdot \|\theta - \theta'\|_2. \]
Notice that we conduct a normalization on $\{a_k\}$ in the encoder and the decoder. 
\end{lemma}
\begin{proof}[Proof of Lemma \ref{lemma: lipschitz}]
Notice that for $\theta = (\frac{m}{2}a_k, \tb_k, c_k)$ and $\theta'=(\frac{m}{2} u_k, \vv_k, w_k')$, we have
\begin{align*}
|f(\theta; \bx) - f(\theta'; \bx)| &= \left|\sum_{k=1}^m a_k \sigma(\tb_k \cdot \bx + c_k) - \sum_{k=1}^m u_k\sigma(\vv_k \cdot \bx + w_k)\right| \\
&\leqslant \left|\sum_{k=1}^m (a_k-u_k)\cdot \sigma(\tb_k\cdot \bx + c_k) + \sum_{k=1}^m u_k\cdot \left(\sigma(\tb_k \cdot \bx + c_k) - \sigma(\vv_k \cdot \bx + w_k)\right)\right| \\
& \leqslant 2 \|a - u\|_1 + \frac2m \cdot \sum_{k=1}^m \left(|(\tb_k-\vv_k)\cdot \bx| + |c_k-w_k|\right) \\
& \leqslant \frac{4}{m} \left\|\frac{m}{2}a - \frac{m}{2}u\right\|_1 + \frac2m\sum_{k=1}^m\|\tb_k-\vv_k\|_1 + \frac2m \|c-w\|_1  \\
& \leqslant \frac{4}{\sqrt{m}} \left\|\frac{m}{2}a - \frac{m}{2}u\right\|_2 + \frac{2\sqrt{d}}{m} \sum_{k=1}^m \|\tb_k-\vv_k\|_2 + \frac{2}{\sqrt{m}} \|c-w\|_2 \\
& \leqslant \left(\frac{16}{m} + m\cdot \frac{4d}{m^2} + \frac{4}{m}\right)^{1/2}\cdot \left(\left\|\frac{m}{2}a - \frac{m}{2}u\right\|_2^2 + \sum_{k=1}^m \|\tb_k-\vv_k\|_2^2 + \|c-w\|_2^2\right)^{1/2} \\
& = \frac{\sqrt{20+4d}}{\sqrt{m}} \cdot \|\theta-\theta'\|_2 \lesssim \sqrt{\frac{d}{m}}\cdot \|\theta-\theta'\|_2. 
\end{align*}
The inequality above holds for $\forall \bx\in \Omega = [0,1]^d$, which leads to
\[\|f(\theta; \cdot) - f(\theta'; \cdot)\|_2 \leqslant \|f(\theta; \cdot) - f(\theta'; \cdot)\|_{\infty} \lesssim \sqrt{\frac{d}{m}}\cdot \|\theta-\theta'\|_2,\]
and it comes to our conclusion. 
\end{proof}

This lemma concludes that $\mathrm{Lip}(G) \lesssim \sqrt{d/m}$ in ~\eqref{eqn: framework}. Finally, the latent dimension $D=m(2+d)$ and the oracle loss
\[W_1(p, \hat{p})\lesssim \varepsilon(D,n)=n^{-1/D}=\exp\left(-\frac{\log n}{m(2+d)}\right). \]
To sum up, it holds that
\[W_1(P, \widehat{P}) \lesssim \frac{4}{\sqrt{m}} + \frac{\sqrt{d}}{\sqrt{m}}\exp\left(-\frac{\log n}{m(2+d)}\right).\]
When choosing the optimal network width $ m = \frac{2\log n}{(2+d)\log d}$, we obtain the optimal rate: 
\[W_1(P, \widehat{P}) \lesssim \sqrt{\frac{d \log d}{\log n}}. \]

\subsection{\color{black} Lower bound: Proof of Theorem \ref{thm: lower}}
\label{app:lower}

{\color{black}
\paragraph{Step 1: Bi-Lipschitz embedding of a Euclidean ball into RKHS Ball $\mathcal B$} ~\\
For a fixed dimension $d\geqslant 1$, define the following linear map $T_d:\bR^d\rightarrow \mH$ by 
\[S_d(t) := \sum_{i=1}^d \sqrt{\mu_i}\cdot t_i \phi_i, \]
where $\{\mu_i\}$ and $\{\phi_i\}$ are the eigenvalues and eigen-functions of RKHS $\mH$, respectively. For vector $t\in\bR^d$, we have
\[\left\|T_d(t)\right\|_{\mH}^2=\sum_{i=1}^d \frac{(\sqrt{\mu_i}\cdot t_i)^2}{\mu_i} = \sum_{i=1}^d t_i^2 = \|t\|_2^2. \]
Hence $T_d: t\in B_d(1) \mapsto T_d(t) \in \mathcal B$, where $B_d(1) := \{t\in\bR^d~:~\|t\|_2\leqslant 1\}$ is a $d$-dimensional Euclidean ball. Moreover, for all $s, t\in \bR^d$, we have
\[\|T_d(t)-T_d(s)\|^2_{L_2(\nu)} = \sum_{i=1}^d \mu_i (t_i-s_i)^2 ~~\Rightarrow~~ \sqrt{\min_{i\in [d]} \mu_i}\cdot \|t-s\|_2 \leqslant \|T_d(t)-T_d(s)\|_{L_2(\nu)} \leqslant \sqrt{\max_{i\in [d]} \mu_i} \cdot\|t-s\|_2.\]
Since $\mu_1 \asymp 1, \mu_d \asymp d^{-2\beta}$, the lower Lipschitz factor is 
\[L_d := \sqrt{\mu_d} \geqslant \sqrt{c}\cdot d^{-\beta}. \]
Now, we introduce the pushforward class. Let $\mathcal Q_d$ be the set of probability measures supported on the $d$-dimensional unit Euclidean ball $B_d(1)$. Consider its image
\[\mathcal P_d := \{T_d \# Q~:~Q\in \mathcal Q_d\} \subseteq \{P~:~\mathrm{supp}(P)\subseteq \mathcal B\}.\]
\paragraph{Step 2: Transferring Wasserstein distance through the embedding} ~\\
For any $Q, Q' \in \mathcal Q_d$, by the definition of $W_1$ and the lower Lipschitz bound, 
\[W_1\left(T_d \# Q, T_d \# Q'\right) \geqslant L_d \cdot W_1^{\bR^d} (Q, Q'),\]
where $W_1^{\bR^d}$ is the Wasserstein-1 metric in $\bR^d$ with Euclidean ground norm. Therefore, we can apply estimator pullback technique and use the statistical lower bound for density estimation inside $d$-dimensional unit ball (Euclidean space) to cap the statistical lower bound for density estimation inside an RKHS ball (function space). 

Given any estimator $\widehat{P}_n = \widehat{P}_n(X_{1:n})$ taking values in probability measures on $\mathcal B$, define a corresponding $\bR^d$-estimator $\widehat{Q}_n$ on data $Y_i\in B_d(1)$ by
\[\widehat{Q}_n (Y_{1:n}) \in \arg\min_{Q\in \mathcal Q_d} W_1\left(T_d\# Q, \widehat{P}_n\left(T_d(Y_1), T_d(Y_2), \ldots, T_d(Y_n)\right)\right). \]
By the optimality of the argmin and the inequality above, we can conclude that
\[W_1\left(T_d\# Q, \widehat{P}_n (T_d(Y_{1:n}))\right) \geqslant L_d \cdot W_1^{\bR^d}(Q, \widehat{Q}_n(Y_{1:n})). \]
Taking expectation under $Y_i\sim Q$ and taking supremum over $Q\in \mathcal Q_d$, we obtain the risk domination 
\[\sup_{P\in \mathcal P_d} \bE\left[W_1(P, \widehat{P}_n)\right] \geqslant L_d \cdot \sup_{Q\in \mathcal Q_d} \bE \left[W_1^{\bR^d} (Q, \widehat{Q}_n)\right].\]
Finally, we take the infimum over all original estimators $\widehat{P}_n$ (which induces taking infimum over all such $\widehat{Q}_n$) yields
\[\inf_{\widehat{P}_n} \sup_{P\in \mathcal P_d} \bE \left[ W_1(P, \widehat{P}_n) \right] \geqslant L_d \cdot \left(\inf_{\widehat{Q}_n}\sup_{Q\in \mathcal Q_d} \bE [W_1^{\bR^d}(Q, \widehat{Q}_n)]\right) := R_n^{(d)}. \]

\paragraph{Step 3: A $d$-dimensional Wasserstein lower bound on the Euclidean ball} ~\\
For each fixed $d\geqslant 1$, a standard conclusion of the classical curse-of-dimensionality for Wasserstein estimation tells us that, on a bounded set of $\bR^d$, the minimax $W_1$ risk scales as $R_n^{(d)} \gtrsim n^{-1/d}$. Therefore,
\[\inf_{\widehat{P}_n} \sup_{P\in \mathcal P_d} \bE \left[W_1(P, \widehat{P}_n)\right] \geqslant L_d \cdot R_n^{(d)} \gtrsim d^{-\beta}\cdot n^{-1/d}.  \]

\paragraph{Step 4: Optimize the bound over dimension $d$} ~\\
For any $d \geqslant 1$, since $\mathcal P_d \subset \{P~:~\mathrm{supp}(P) \subseteq \mathcal B\}$, we have
\[\mathcal R_n := \inf_{\widehat{P}_n} \sup_{P:~\mathrm{supp}(P)\subseteq \mathcal B} \geqslant \inf_{\widehat{P}_n} \sup_{P\in \mathcal P_d} \bE \left[W_1(P, \widehat{P}_n)\right] \gtrsim d^{-\beta}\cdot n^{-1/d}. \]
Now we choose an optimal dimension $d$ to maximize $d^{-\beta}\cdot n^{-1/d}$. By basic calculus, it is obvious that the maximizer $d^* = \frac{\log n}{\beta}$, where $d^{-\beta}\cdot n^{-1/d} = (\beta / e)^{\beta}\cdot \left(\frac{1}{\log n}\right)^{\beta}$. To sum up, we have
\[\mathcal R_n := \inf_{\widehat{P}_n} \sup_{P:~\mathrm{supp}(P)\subseteq \mathcal B} \gtrsim \left(\frac{1}{\log n}\right)^{\beta},\]
which comes to our conclusion.
}

\clearpage
\section{Experimental details}

\label{appendix: experiments}

\subsection{Baselines}
\label{app:baselines}

{\color{black}
In our experiments, all baseline models are trained using the same procedure described in the Results section. 
Specifically, each model is trained for $10^5$ iterations using the AdamW optimizer with a weight decay of $10^{-5}$ and a batch size of 16. 
The learning rate schedule includes a warm-up of 2{,}000 steps from zero to the base learning rate. 
For neural operator baselines, we use a base learning rate of $10^{-3}$ with exponential decay every 2{,}000 steps at a rate of 0.9. 
For diffusion-based models, we use a base learning rate of $10^{-4}$ with the same decay schedule, and apply exponential moving average (EMA) updates to stabilize training. 
For randomly downsampled inputs, we interpolate them back to the original resolution before processing.}

\paragraph{Fourier Neural Operator (FNO).} 
FNO architecture leverages Fourier transform to efficiently learn mappings between function spaces, making it particularly effective for modeling PDEs and physical systems.
We employ an FNO model with four layers, where the number of channels is 32, and the Fourier modes is 32. The GeLU activation function is employed, and no normalization scheme is applied.

\paragraph{Vision Transformer (ViT).}
ViT extends the transformer architecture to visual tasks by utilizing self-attention mechanisms that capture global dependencies across input fields, offering superior performance on a range of spatial prediction problems.
We adopt a ViT architecture that processes inputs by splitting them into patches of size 16. The network consists of 8 transformer blocks, each with an embedding dimension of 256, an MLP ratio of 2, and 8 attention heads. 

\paragraph{UNet.}
UNet implements a symmetric encoder-decoder framework with skip connections that preserves multi-resolution features, enabling effective semantic segmentation and dense prediction tasks. We implement a conventional UNet architecture \cite{ronneberger2015u} with group normalization. The model uses an embedding dimension of 64 for its convolutional layers. We use GeLU activations and group normalization.

{\color{black}
\paragraph{Continuous Vision Transformer (CViT) \cite{wang2024cvit}.} A neural field architecture that integrates a ViT encoder with trainable grid and coordinate embeddings for continuous, resolution-agnostic querying. In our implementation, the encoder operates on $16\times16$ patches with an embedding dimension of 256, depth 8, and 8 attention heads (MLP ratio 2). 
The decoder adopts a similar configuration with an embedding dimension of 256, depth 4, and 8 attention heads, followed by two MLP layers to obtain the final output.

\paragraph{CNextU-Net \cite{ohana2024well}.} A modernized U-Net backbone built upon ConvNeXt blocks \cite{woo2023convnext}, providing a high-capacity convolutional baseline competitive with transformer-based architectures for spatiotemporal field regression. our model employs a four-stage encoder–decoder hierarchy with two ConvNeXt blocks per stage and two additional blocks at the bottleneck, initialized with 32 feature channels and doubling features at each downsampling level.

\paragraph{AviT \cite{mccabe2023multiple}.} An attention-based ViT variant that sequentially attends over spatial and temporal dimensions, enabling efficient modeling of dynamic physical systems.  Our implementation follows the base model described in Ref.~\cite{mccabe2023multiple}, consisting of 8 AViT blocks with hidden dimension 512, 8 attention heads, and grouped axial attention across spatial partitions.

\paragraph{DPOT \cite{hao2024dpot}.} A PDE foundation operator trained via autoregressive denoising and Fourier attention, designed to generalize across diverse PDE systems and boundary conditions.  In our implementation, DPOT processes inputs of size $256\times256$ with a patch size of 16, embedding dimension 512 and an MLP ratio of 2. The core architecture consists of a Fourier-attention transformer stack with 8 layers, interleaved with 4 Fourier mixer blocks (via AFNO) for spatial-frequency mixing.

\paragraph{Diffusion Posterior Sampling (DPS) \cite{chung2023diffusion}.} 
An inverse solver for general operators $\mathcal{A}$ that augments the reverse diffusion process with a likelihood-score guidance term $\nabla_x \log p_\sigma(y\,|\,\mathcal{A}(x))$ to approximate posterior sampling. 
We adopt the original denoising diffusion probabilistic model (DDPM) backbone with four encoder blocks and feature depths $(32, 64, 128, 256)$, each comprising two residual blocks and group normalization (8 groups). 
Attention is applied in the last two stages with eight heads, and the model uses an embedding width of 256 and one middle residual block. 
Sampling is performed over 1{,}000 diffusion steps using a cosine noise schedule and a tuned guidance scale of $\lambda = 1000$, selected for best reconstruction accuracy.

\paragraph{DiffusionPDE \cite{huang2024diffusionpde}.} 
A conditional diffusion model that jointly learns the distribution of PDE coefficients and solutions, guided by both observation consistency and PDE residual constraints during denoising. 
We follow the official implementation with a four-stage U-Net backbone using feature channels $(64, 128, 256, 512)$, two residual blocks per stage, and self-attention in the bottleneck layer. 
Training is conducted for 1{,}000 diffusion steps with a cosine noise schedule and dual guidance weights tuned to balance data fidelity and physics consistency, selected for optimal reconstruction accuracy.

\paragraph{Cross-Attention Diffusion (Cross-Attn) \cite{zhuang2025spatially}.} 
A diffusion framework that employs cross-attention to reconstruct spatially distributed physical fields from sparse or irregular measurements. 
Following the original implementation, we use the aforementioned U-Net as the backbone. 
For the conditioning, measurement tokens are obtained through patch embeddings and processed by 4 transformer layers with an embedding dimension of 256, MLP ratio 2, and 8 attention heads. 
Cross-attention modules are integrated into each U-Net residual block to encode measurement-conditioned context and guide the denoising process. 
}

\subsection{Kolmogorov flow}

\paragraph{Data generation.} 
We utilize the dataset generated by Li et al.~\cite{li2021fourier}, solved using a pseudo-spectral solver. The initial condition $\omega_0(x)$ is sampled from a Gaussian random field, $\mathcal{N}\left(0,73 / 2(-\Delta+49 I)^{-5 / 2}\right)$. The simulation employs a $2048 \times 2048$ uniform grid, with 40 sequences spanning 10 seconds each $(T=10)$. The data is downsampled to a $256 \times 256$ spatial grid with a fixed temporal resolution of $\Delta t=$ $1 / 32$ seconds, yielding a dataset of 40 sequences, each containing 320 frames.

\begin{figure}[htbp]
    \centering
    \includegraphics[width=1.0\linewidth]{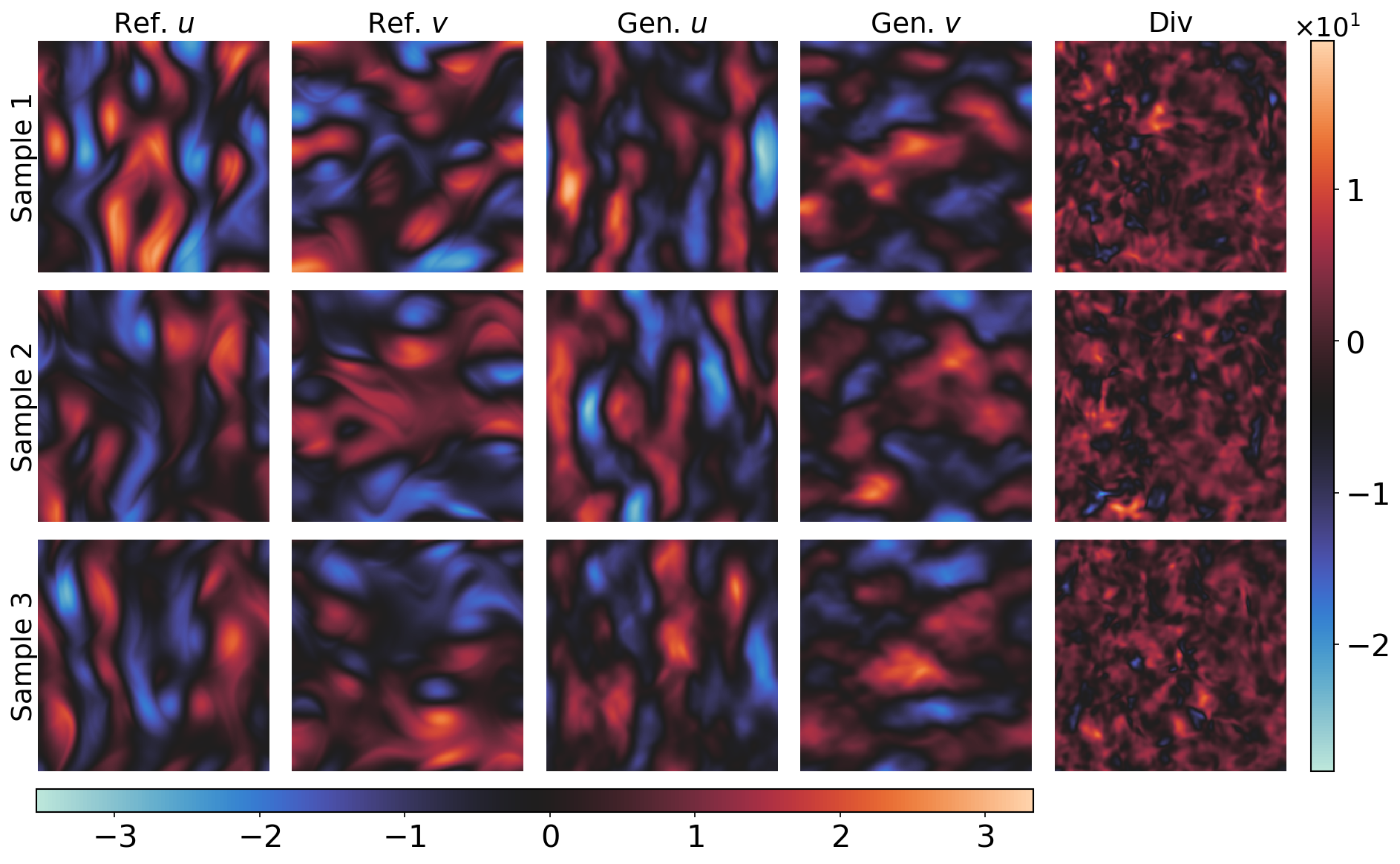}
    \caption{\textbf{Kolmogorov flow.} Unconditional generation without enforced divergence-free constraint. Examples of three generated velocity fields and their corresponding divergence fields.}
    \label{fig:kf_non_div_samples}
\end{figure}

\begin{figure}[htbp]
    \centering
    \includegraphics[width=1.0\linewidth]{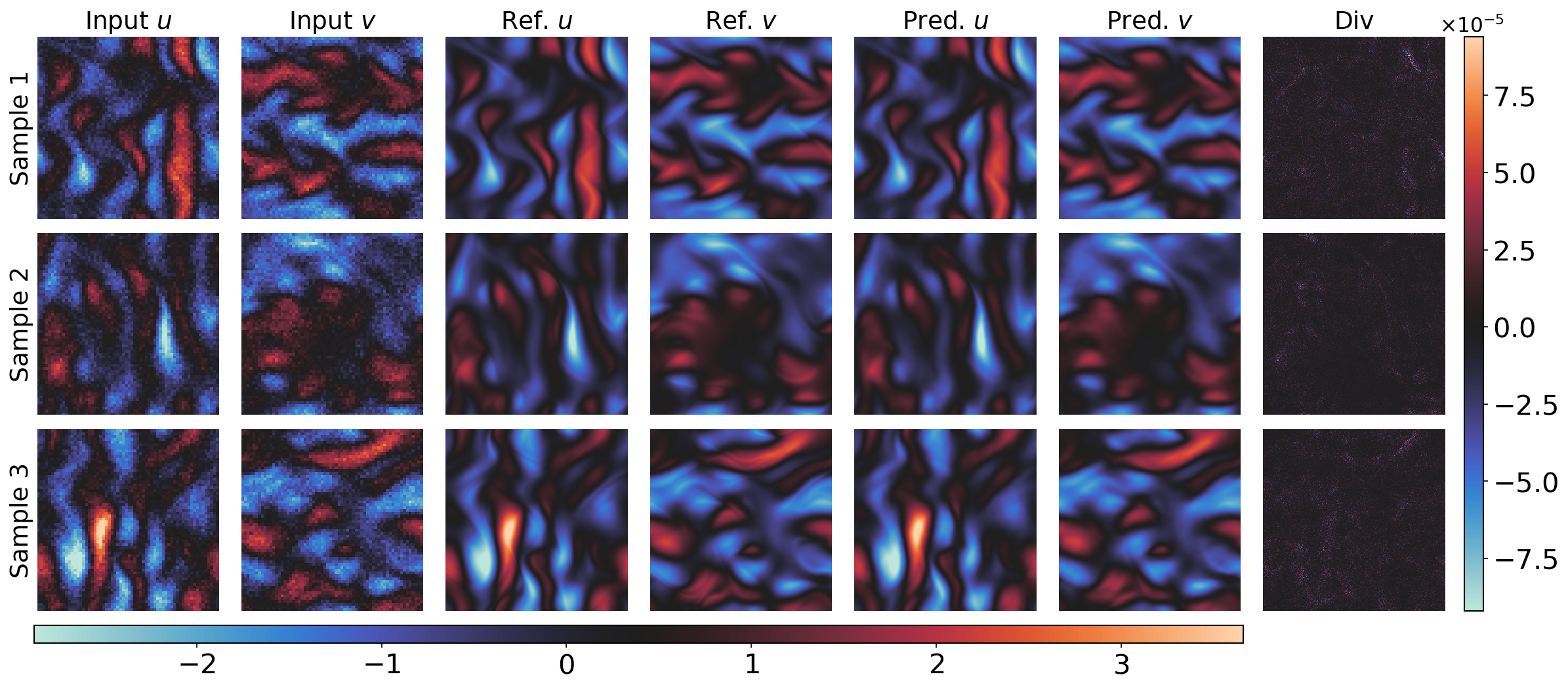}
    \caption{\textbf{Kolmogorov flow.} Examples of the model predictions for three sample solutions. Each row shows a different sample, displaying from left to right: downsampled input velocity components ($u$, $v$), reference high-resolution velocity fields, predicted velocity fields, and the predicted divergence field. The model successfully reconstructs sharp flow features from the coarse input while strictly maintaining the divergence-free condition, as evidenced by the negligible values in the divergence fields (rightmost column).}
    \label{fig:kf_samples}
\end{figure}

\begin{figure}[htbp]
    \centering
    \includegraphics[width=1.0\linewidth]{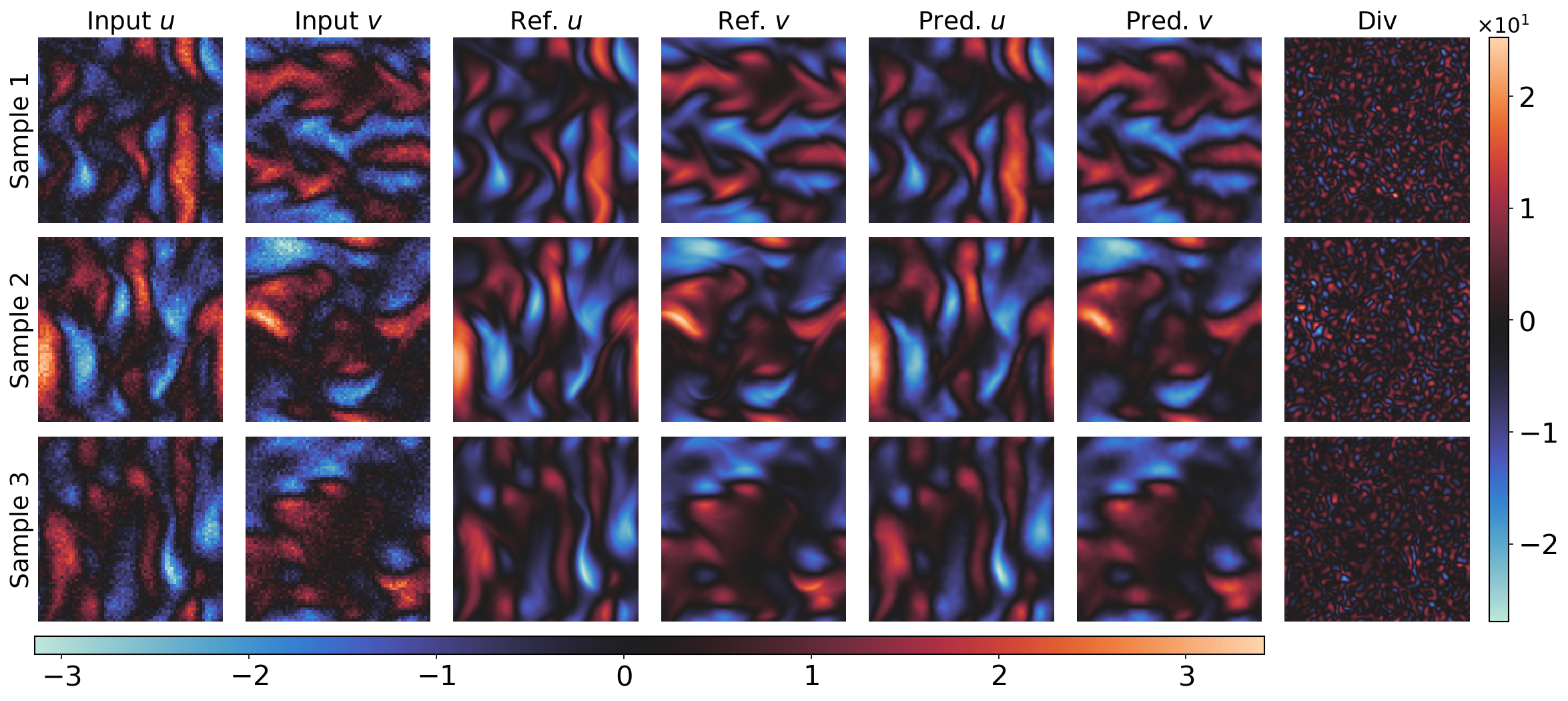}
    \caption{\textbf{Kolmogorov flow.} Examples of the model predictions for three sample solutions. Each row shows a different sample with: downsampled input velocity components ($u$, $v$), reference high-resolution fields, predicted velocity fields, and the predicted divergence. While the model achieves basic upsampling of the velocity fields, the reconstructions exhibit noticeable blurring compared to the sharp features in the reference solutions. The large values in the divergence fields (rightmost column) indicate that the incompressibility constraint is not enforced.}
    \label{fig:kf_no_div_free_samples}
\end{figure}

\subsection{Burgers' equation}

\paragraph{Data generation.}
Following Wang et al.~\cite{wang2021learning}, we generate our dataset by sampling 4,000 initial conditions from a GRF with covariance operator $\mathcal{N}(0,25^2(-\Delta+5^2 I)^{-4})$. For each initial condition, we solve the Burgers' equation in ~\eqref{eq:burgers} using spectral methods with periodic boundary conditions. The numerical solution is computed using the Chebfun package \cite{driscoll2014chebfun} with a spectral Fourier discretization and the fourth-order ETDRK4 time-stepping scheme \cite{cox2002exponential}, using a time-step size of $10^{-4}$. We record solution snapshots every $\Delta t=0.005$ up to $t=1$, resulting in 201 temporal snapshots for each realization on a uniform $200 \times 201$ spatio-temporal grid.

\begin{figure}[htbp]
    \centering
    \includegraphics[width=0.9\linewidth]{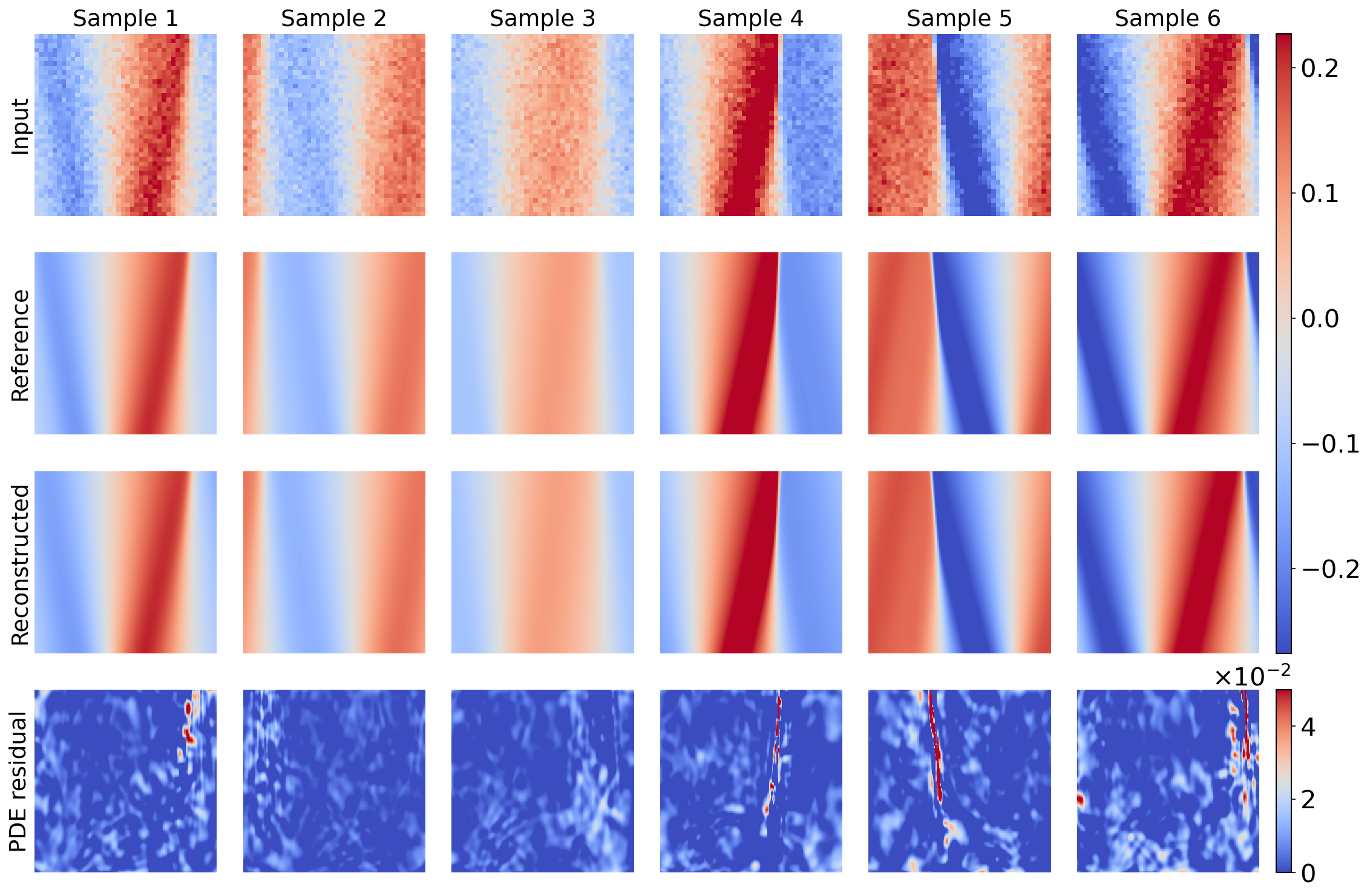}
    \caption{\textbf{Burgers' equation.} Representative generated PDE solutions and the corresponding PDE residuals, given the coarse measurements in $50 \times 50$ space-time resolution. The model is trained with incorporated PDE residual constraints.}
    \label{fig:burger_pde_samples}
\end{figure}

\begin{figure}
    \centering
    \includegraphics[width=0.9\linewidth]{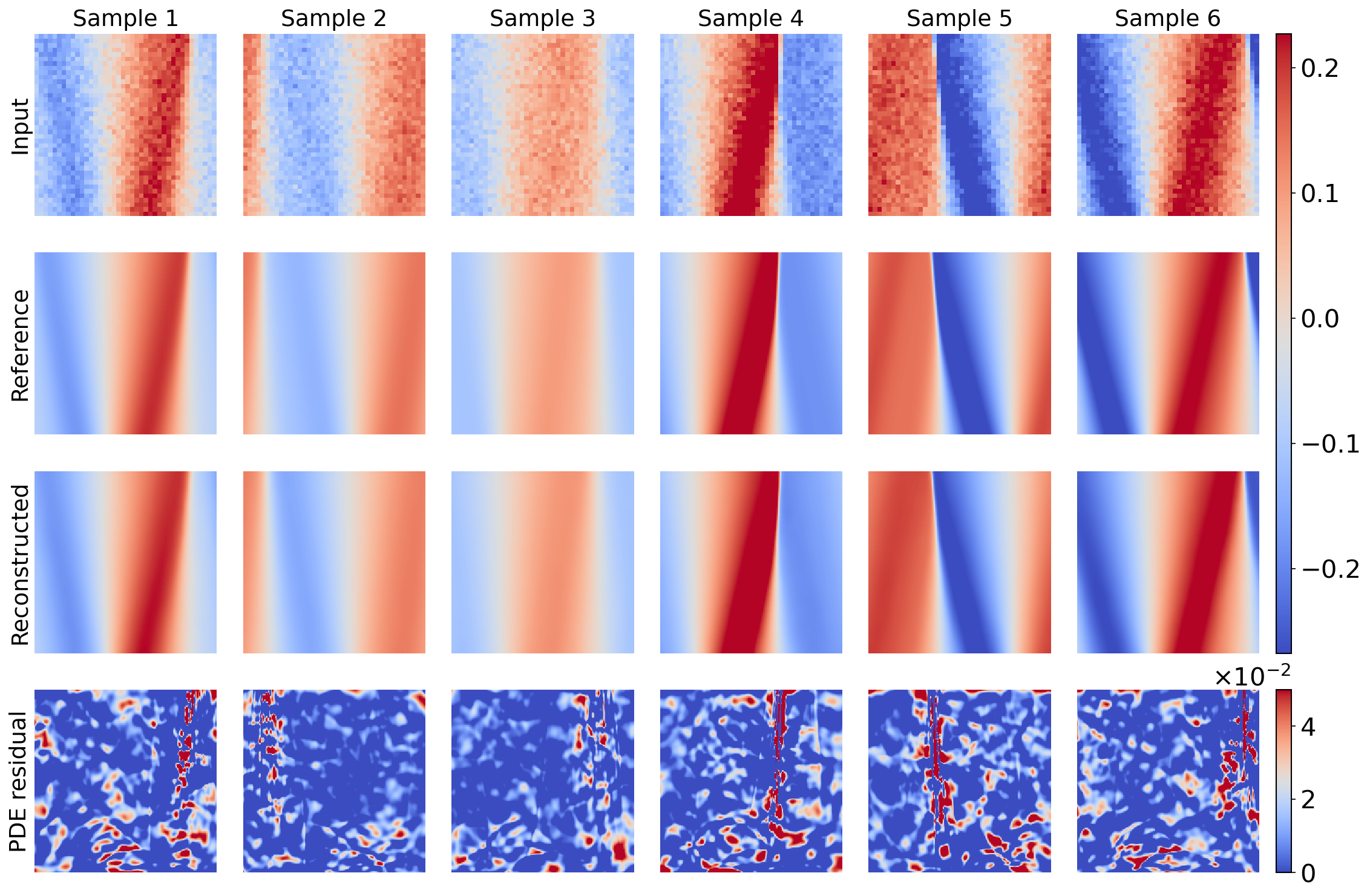}
    \caption{\textbf{Burgers' equation.} Representative generated PDE solutions and the corresponding PDE residuals, given the coarse measurements in $50 \times 50$ space-time resolution. The model is trained without incorporating PDE residual constraints.} 
    \label{fig:burgers_no_pde_samples}
\end{figure}

\subsection{Linear elasticity}

\paragraph{Data generation.}
The RVE domain $\Omega=[0, 1]^2$ is considered as a fiber-reinforced composite microstructure, where circular fibers are embedded within the matrix in a hexagonal lattice arrangement. The material properties including $E(\mathbf{x})$ and $\nu(\mathbf{x})$ are homogeneous inside the disjointed matrix or fiber phases:
\begin{equation*}
E(\mathbf{x}), \nu(\mathbf{x})= \begin{cases}
            E_{f}, \nu_{f}, & \text { if } \mathbf{x} \in \Omega_{f}, \\
            E_{m}, \nu_{m}, & \text { if } \mathbf{x} \in \Omega_{m},
\end{cases}
\end{equation*}
where ${\Omega}_f$ and ${\Omega}_m$ represent fiber and matrix phases, respectively.
Our dataset comprises 7,500 unique RVEs, each discretized on a uniform $256 \times 256$ grid. For all RVEs, the matrix material properties are fixed with $E_m = 1$ (GPa) and $\nu_{m} = \nu_{f} = 0.35$. The volume fraction $vof$ and the Young's modulus $E_f$ of fiber are randomly generated through the Latin hypercube sampling (LHS) design method, ensuring that $vof \in [0.1,0.65]$ and $E_f \in [10^{-6},10^{6}]$. The $A_{i j k l}(\mathbf{x})$ of each RVE is computed by solving ~\eqref{setup:dotLS} subjected to ~\eqref{setup:macroBC} using a Fast Fourier Transform (FFT)-based homogenization method \cite{Moulinec1998FFT}. A detailed and comprehensive implementation of generating $A_{i j k l}(\mathbf{x})$ can be found in Ref.~\cite{WangLiuMicrometer}. Given that the diagonal components $A_{11}$, $A_{22}$, and $A_{33}$ of the strain concentration tensor dominate the mechanical response, these components are exclusively used to train our functional autoencoder.

\begin{figure}[htbp]
    \centering
    \includegraphics[width=0.6\linewidth]{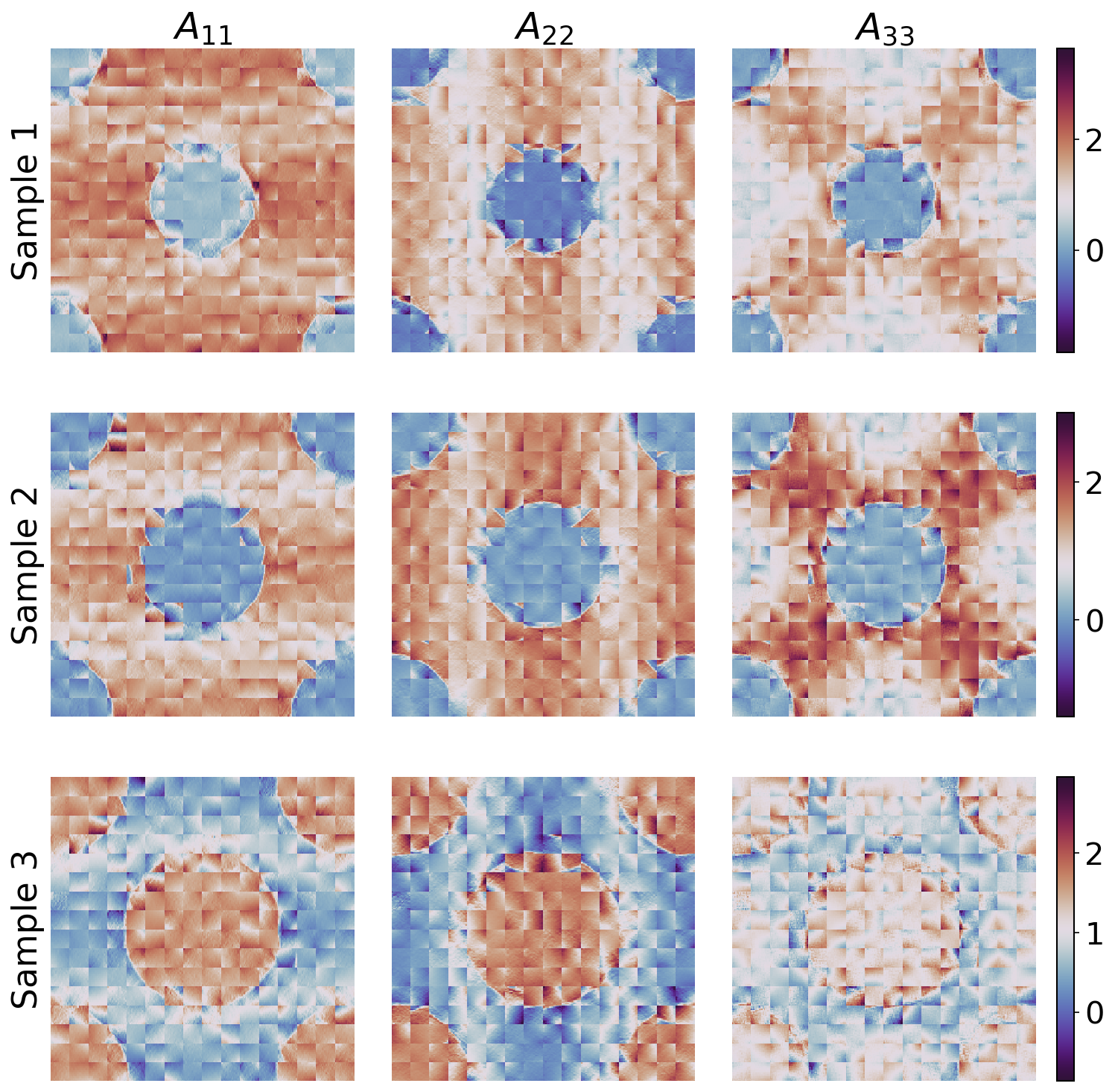}
    \caption{\textbf{Linear elasticity.} Generative samples obtained by training a discrete ViT autoencoder and latent diffusion model using the pipeline in Fig.~\ref{fig:pipeline}. Noticeable discontinuities appear in the generated outputs, manifesting as distinct patch artifacts. }
    \label{fig:elasticity_discrete}
\end{figure}

\subsection{Turbulence mass transfer}

\paragraph{Data generation.}
Various topologies are randomly generated and evenly assigned them to 9 different inlet velocities $(v \in\{0.1,0.2, \ldots, 0.9\} \mathrm{m} / \mathrm{s})$. The mechanistic model is used to obtain the concentration and pressure distributions, corresponding to the given $v$ and $\gamma(x, y)$. To facilitate the training of neural networks, the total pressure $p$, including static pressure and dynamic pressure, is scaled to a uniform range, where $P=\frac{p}{2000}+0.26$. To ensure the effectiveness of the generated topology, the following solid volume constraint is set:
\begin{align*}
    \iint(1-\gamma) d V \geq \alpha V_{\Omega},
\end{align*}
where $\alpha=0.1$ is the minimum solid volume fraction in the design domain, and $V_{\Omega}$ is the volume of the design domain. We have 900 samples for training and 55 for testing.